\documentclass{article}

\usepackage{PRIMEarxiv}
\usepackage{natbib}
\usepackage[utf8]{inputenc} 
\usepackage[T1]{fontenc}    
\usepackage{hyperref}       
\usepackage{url}            
\usepackage{booktabs}       
\usepackage{amsfonts}       
\usepackage{nicefrac}       
\usepackage{microtype}      
\usepackage{lipsum}
\usepackage{fancyhdr}       
\usepackage{graphicx}       
\graphicspath{{media/}}     
\usepackage{multirow}
\usepackage{subfigure}

\usepackage{graphicx}
\usepackage{subcaption}
\usepackage[utf8]{inputenc}
\usepackage{xcolor}
\usepackage{amsmath}
\usepackage{tcolorbox}
\tcbuselibrary{skins, breakable} 

\definecolor{boxbg}{HTML}{F8F9FA}
\definecolor{boxborder}{HTML}{D1D5DB}
\definecolor{headerbg}{HTML}{4B5563}

\usepackage{tcolorbox}
\tcbuselibrary{skins}

\definecolor{promptborder}{HTML}{D1D5DB}
\definecolor{promptheader}{HTML}{374151}

\tcbset{
    promptbox/.style={
        enhanced,
        colback=gray!5,
        colframe=promptborder,
        coltitle=white,
        colbacktitle=promptheader,
        fonttitle=\bfseries\small,
        boxrule=0.6pt,
        arc=3pt,
        left=8pt,
        right=8pt,
        top=6pt,
        bottom=6pt
    }
}

\title{Prompt Lottery in LLM-as-a-Judge: Measuring and Mitigating Sensitivity to Syntactic Rephrasing

}

\author{
  Ilija Subasic, Brett Levert, Andrew Rabinovich, Sajjad Pakdamansavoji, Zhao Chen \\
  Upwork Inc \\
  \texttt{firstname.lastname@\{cloud.\}upwork.com} \\
}

\begin{document}

\maketitle

\begin{abstract}
LLM-as-a-Judge (LLMaJ) has become a standard paradigm for evaluating large language models, yet most approaches implicitly assume that a single judge prompt functions as a stable evaluation instrument. Prior work has primarily explored deliberate prompt modifications such as scoring rubrics, role descriptions, or chain-of-thought reasoning. In contrast, we investigate whether validated prompt variants intended to preserve evaluation semantics while differing in surface realization (\textit{paraphrastic rephrasings}) yield consistent judgments under a fixed evaluation task. We introduce \textit{Prompt Lottery}, a phenomenon in which prompt reformulations intended to preserve semantic content produce different LLMaJ judgments. Across a large-scale study spanning diverse evaluation tasks, judge families, and model backbones, we find that paraphrastic variation can alter evaluation outcomes and, in some settings, reverse relative model rankings. We conducted a user study and show that the variability of LLMaJ exceeds both the variability observed among human annotators in our study and the generally accepted variability in human-subject studies, raising concerns about the stability of single-prompt evaluation. Finally, we show that multi-prompt aggregation substantially improves robustness by emphasizing judgments that remain consistent across prompt variants.
\end{abstract}

\section{Introduction}

Since their introduction as a reinforcement learning technique in Constitutional AI \cite{bai2022constitutionalaiharmlessnessai}, LLM-as-a-judge (LLMaJ) use has become central to evaluating Large Language Models. The basic premise is that given an (LLM-generated) input $I$, a set of instructions $P$ (prompt), and sometimes a reference answer, a back-model $M$ produces a judgment score as a proxy for the quality of $I$. Early work by \cite{zheng2023judging} reported high agreement between human and model judgments in controlled settings, establishing LLMaJ as a standard evaluation paradigm across multiple tasks \cite{g-eval}. Recent surveys summarize the field along dimensions such as target attributes, evaluation protocols, human alignment, and bias mitigation \cite{gu2024llm, gao2024llm, chang2024survey}. Despite this progress, most research focuses on how judgments are produced (e.g., scoring schemes, multi-agent setups, or aggregation strategies), and less attention has been paid to the initial formulation of the judge prompt, the core instruction that defines the evaluation behavior regardless of any optimization or enhancement.

Despite the known sensitivity of LLMs to prompt phrasing \cite{liu2023pre}, 
current research focuses on optimizing LLMaJ through rubric refinement, bias mitigation, or multi-agent aggregation \cite{gu2024justice, chan2023chateval, mizrahi2024stateartmultipromptllm, pmlr-v267-zhuge25a} and not the semantics of the evaluation instructions. We demonstrate that this foundation may lack empirical grounding. Unlike prior work on prompt enhancement, which alters task instructions to improve performance, we isolate variance arising from paraphrastic prompt variation. We term this the Prompt Lottery: the phenomenon whereby evaluation outcomes depend on prompt formulation choices that, while intended to be semantically equivalent, produce systematically different (though unpredictable to practitioners) judgments. 

\begin{tcolorbox}[promptbox, title=Variant A ($P_1$)]
\small\ttfamily\raggedright
\begin{obeylines}
\{dialogue\} 

Above is a dialogue context and the corresponding response.

Question: Is the response understandable?

\#\#\# 
Your Answer (respond with ONLY 'Yes' or 'No')
\end{obeylines}
\end{tcolorbox}

\vspace{0.3cm}

\begin{tcolorbox}[promptbox, title=Variant B ($P_2$)]
\small\ttfamily\raggedright
\begin{obeylines}
\{dialogue\} 

Is the response in the above conversation clear and easy to comprehend? 

\#\#\# 

Answer with 'Yes' or 'No'.
\end{obeylines}
\end{tcolorbox}

An LLMaJ's sensitivity to prompt variation is a strong indicator of evaluator robustness. If a researcher concludes that a new model outperforms a baseline using a single prompt, but an equally valid rephrasing reverses that conclusion, it reveals single-prompt evaluation to be fragile and misleading. Paraphrastic robustness can be viewed as a form of evaluation invariance: prompt variants intended to preserve semantic meaning should yield stable judgments. Violations of this property suggest that evaluation outcomes may depend partly on prompt formulation rather than only on the evaluated response. Prior work has quantified these effects in human annotation and developed strategies to ensure judgments reflect task semantics rather than syntactic framing \cite{artstein2008intercoder}. These mitigation strategies are grounded in a core requirement: semantically equivalent instructions should yield consistent judgments—
a requirement we term paraphrastic robustness, central to principled evaluation. If LLMaJ pipelines are to replace human evaluators, they must satisfy this same invariance property.

In this paper we answer several questions connected with LLMaJ robustness to paraphrastic variations based on an array of experiments
over 825 conditions spanning 12 datasets and 6 different judge families. We first investigate whether agreement between LLMaJ holds when prompts are changed. We 
show that LLMaJ evaluators exhibit larger variance than observed in human annotators, and beyond the generally acceptable level for human user studies,
degrading from near-perfect to fair agreement. We then ask whether this degradation is important and demonstrate that it is not inconsequential: it leads to lower alignment with human judgments and can lead to model rank reversal in more objective tasksß. This prompts us to investigate how to improve the robustness of LLMaJ
and we introduce an efficient aggregation strategy that restores reliability while reducing inference costs by 60\% compared with naive aggregation. This selective aggregation approach achieves 86\% of full robustness at more than half of the compute savings ($\sim 2\times$ instead of $5\times$ increase relative to not aggregating).
 
\section{Related Work}

Quality human annotation relies on abstracting from surface-level variations to capture underlying semantic intent. This objective is rooted in the distinction between linguistic forms and their core semantic content \citep{chomsky1965aspects}, and is enacted through rigorous annotation protocols designed to maximize inter-coder agreement \citep{artstein-poesio-2008-survey, pavlick2019paraphrase}. As human judgments can be susceptible to the Framing Effect in heuristic decision-making \citep{tversky1981framing}, controlled evaluations utilize coding instructions to remove these biases and ensure reliability \citep{krippendorff2018content}. If LLMaJ pipelines are to replace human evaluators, they must satisfy this same invariance property. A more detailed theoretical analysis of formal semantics and an exploration of why invariance should or should not hold for LLMs are provided in Appendix~\ref{app:semantic_invariance}.

LLMaJ emerged as time and resource saving approach, and is now widely used for both reference-based (ground truth matching) and reference-free evaluation (free observation) tasks, often employing pairwise ranking or rubric-based scoring \citep{zheng2023judging, g-eval}. Modern frameworks incorporate agentic reasoning or multi-step pipelines to approximate human judgment processes \citep{pmlr-v267-zhuge25a}. Despite success, their sensitivity to prompt design raises questions about their stability as evaluation instruments. Recent surveys structure LLMaJ research along four axes: (i) target attributes, (ii) evaluation protocols, (iii) alignment with human judges, and (iv) bias analysis \citep{gu2024llm, gao2024llm, chang2024survey}. While these emphasize scaling and aggregation strategies, they largely overlook the effect of initial evaluator prompt formulation. Our work complements this line by focusing on stability under syntactic variation rather than optimization under aggregation. 

Zheng et al. (\citep{zheng2023judging}) show that targeted prompt modifications, such as a judge persona, score delimiters, and chain-of-thought reasoning, improve human alignment and reduce position and verbosity biases. Beyond this, several works reduce judge bias through aggregation: model ensembles combine multiple back-models \citep{williams20250humanvsai}, while \citep{ape2025} generates augmented instructions and verifies consistency using a single model. Hierarchical tie-breaking resolves disagreements via a secondary model \citep{findings2025}, and \citep{zhu2026cyclicjudgemitigatingjudgebias} rotates models over a fixed prompt to cancel systematic biases. \citep{haldar2025rating} have also identified substantial self-inconsistency in LLM judges across repeated evaluations, even under fixed prompts. Closest settings to ours are \citep{mizrahi2024stateartmultipromptllm} who advocate multi-prompt evaluation using format perturbations (e.g., quotes, line breaks) on fixed instructions, \citep{deng2026rubric} extend this to rubric scoring with synonym and adversarial variations, and \citep{chiang-lee-2023-large} who measure human rating averages across instructions. In contrast, we examine \emph{paraphrastic variation}, where semantic equivalence is guaranteed by construction rather than formatting. While \citep{bellibatlu2026judgesense} proposed a reliability score under paraphrastic rephrasings, they only evaluate models relative to each other and do not assess the system-wide robustness of LLMaJs.

\section{Experimental Setup} \label{sec:judges}

In total we evaluate 33 configurations (12 datasets, 6 judge families) with 5 backbone models, drawn from both proprietary and open providers, and 5 prompt variants each, yielding 825 experimental conditions, as not all judge families apply to every dataset (the summarization judge is used only for summarization tasks). We evaluate five backbone LLMs: GPT-4o, GPT-oss-120B, Llama 3.3 (70B Instruct), Kimi-K2.5(-Instruct), and Qwen3(-VL-30B-A3B-Instruct). For each judge family, we use the original prompt and four semantically equivalent variations (Section~\ref{sec:rephrasing}), yielding 25 distinct configurations per dataset. \citep{li2024necessity} argue that non-zero temperatures capture fuller evaluative potential; however, our focus is the model's capacity for semantic consistency. We therefore use temperature $T=10^{-5}$ to ensure results reflect core interpretive stability rather than ``wisdom-of-the-crowd'' effects, though GPU kernel non-determinism may remain \citep{he2024defeating}.

We utilize judge families grounded in established evaluation frameworks (Table~\ref{tab:judges_datasets}), encompassing G-Eval style multidimensional rubrics~\citep{g-eval} and specialized metrics. These span binary classification and Likert-scale scoring across objective (Correctness, Groundedness) and subjective (Dialogue Quality, General Response Quality) axes, demonstrating the Prompt Lottery's impact across the wide spectrum of LLM evaluation. We prioritized datasets with accessible codebases and prompt templates to ensure fidelity to original conditions and eliminate confounding from manual reconstruction. The prompts and paraphrastic variants are provided in Appendix~\ref{appendix:prompts}.

\begin{table*}[t]
\centering
\tiny
\setlength{\tabcolsep}{5pt}
\renewcommand{\arraystretch}{1.15}
\captionof{table}{Overview of judge families, evaluation tasks, and datasets.}
\begin{tabular}{p{2.2cm}p{2cm}p{2cm}p{3cm}p{3cm}}
\toprule
\textbf{Judge Family} & \textbf{Dimensions} & \textbf{Scoring Task} & \textbf{Datasets} & \textbf{References} \\
\midrule
Dialogue Quality & \textit{Interesting, Specific, Understandable, Quality} & \textbf{Binary (Yes/No):}  Presence of quality traits in conversation context & PersonaChat, PersonaUSR, DailyDialog, Topical-Chat, ConTurE & \citep{mehri2020usr, Zhang2018PersonaChat, mehri-eskenazi-2020-unsupervised, li2017dailydialog, topical, ghazarian2022wrongyouleveraginguser} \\
\midrule
QA Correctness & \textit{Correctness} & \textbf{Likert 1--5:} Completeness and accuracy vs. ground-truth & HotpotQA, SQuAD v2, NQ-Open, TopiOCQA, Natural Questions & \citep{zheng2023judging, yang2018hotpotqa, rajpurkar2018know, lee2019latent, adlakha2022topiocqa, kwiatkowski2019natural} \\
\midrule
QA Groundedness & \textit{Groundedness} & \textbf{Likert 1--5:} Faithful inference from evidence & HotpotQA, SQuAD v2, NQ-Open, TopiOCQA, Natural Questions & \citep{es2024ragas, yang2018hotpotqa, rajpurkar2018know, lee2019latent, adlakha2022topiocqa, kwiatkowski2019natural} \\
\midrule
General Response Quality & \textit{Helpfulness, Accuracy, Depth} & \textbf{Likert 1--10:} Holistic assessment of open-ended responses & LLM-Arena, PersonaChat, PersonaUSR, DailyDialog, Topical-Chat, ConTurE & \citep{zheng2023judging, Zhang2018PersonaChat, mehri-eskenazi-2020-unsupervised, li2017dailydialog, topical, ghazarian2022wrongyouleveraginguser} \\
\midrule
Coherence, Consistency, Fluency & \textit{Summary Quality} & \textbf{Likert 1--5:} Summarization quality assessment & SummEval & \citep{fabbri2021summeval} \\
\midrule
Semantic Alignment & \textit{Content \& Style Equivalence} & \textbf{Likert 1--5:} Information overlap between reference and prediction & UPHELD & \citep{subasic2026evaluating} \\
\bottomrule
\end{tabular}
\label{tab:judges_datasets}
\end{table*}

We examine whether paraphrastic rephrasing induces significant variance in judgment scores using four metrics: \emph{Overlap} (rate of identical judgments), \emph{Average Pairwise Cohen's $\kappa$}, \emph{Fleiss' $\kappa$}, and \emph{Krippendorff's} $\alpha$ (agreement across all rephrased prompts). We report all metrics but focus our analysis on $\kappa$ scores for direct comparison with prior work on human judgment under instruction variation, while also noting high correlation with Krippendorff's $\alpha$ and Fleiss' $\kappa$.

\subsection{Paraphrastic Prompt Generation and Validation} \label{sec:rephrasing}

We generated four paraphrastic variants using Claude 3 Opus. These were reviewed by human annotators for preservation of intended task meaning and verified using Gemini 1.5 Pro for syntax validity. To characterize the variations qualitatively, we post hoc identified six main, recurring, but not mutually exclusive dimensions of surface variation, including role framing, instruction syntax, contextualization, and output specification (Table~\ref{appendix:prompt_dim}); a within-family permutation test of these dimensions against cross-model Cohen's $\kappa$ (50 judge prompts, 46 experiments, 10{,}000 permutations per contrast, Holm-adjusted) found no dimension that consistently produced higher agreement across models (Appendix~\ref{appendix:prompt_diminsions} contains full details). First, three independent LLMs (Kimi-K2.5, Gemini 1.5 Pro, and GPT-4o) evaluated pairwise prompt similarity, yielding average scores of 81\%, 84\%, and 79\%, respectively. Second, sBERT semantic similarity \citep{bert} averaged 0.93 across all prompt pairs, exceeding a commonly used high-similarity threshold of 0.80. In contrast, lexical overlap remained comparatively low (Jaccard: $0.38 \pm 0.01$). High semantic similarity and low lexical overlap suggest that prompt variants preserved intended evaluation content while differing substantially in wording, consistent with successful paraphrastic rephrasing.

To strengthen these findings, we conducted a human validation study focused on task equivalence rather than surface-level similarity. Each pair of generated prompts (including the original prompt) was shown to three annotators, who rated whether the two prompts require the same evaluation task. Annotators consistently judged the prompts to be highly equivalent, with an average POMP \citep{cohen1999problem} score of 76/100. Inter-annotator agreement was substantial (Cohen's $\kappa = 0.78$; Krippendorff's $\alpha=0.81$), indicating that this perception of equivalence was consistent across raters.

\subsection{Annotator Alignment Study and Expected Agreement Levels}
While humans perceive the instruction prompts we created as equivalent tasks, prior work on annotation guideline rephrasing suggests this may not guarantee identical annotation behavior \citep{benikova2014nosta}. To test this, we conducted a user study in which we recruited 10 annotators (at a fair market rate) via a freelancer agency. We selected annotators with high success rates, indicating professionalism and skill. They were presented with randomized pairs of instructions for scoring answers, the answers, and context (e.g., conversation so far, or a question). The instructions were identical to the paraphrased prompt variants. Annotators provided individual ratings, and we ensured annotators never rated the same data point with different instructions back-to-back. In total we collected annotations on 650 data points, and each annotator sequentially rated examples, with each example annotated by at least three annotators. Based on these annotations, we calculated agreement metrics for humans. On average across prompts and datasets, we observed Overlap, Average Pairwise Cohen's $\kappa$, Fleiss' $\kappa$, and Krippendorff's $\alpha$ (all normalized to 0--1) of $0.78$, $0.71$, $0.80$, and $0.63$, respectively. Findings from previous work are consistent with our study: \citep{benikova2014nosta} found that paraphrasing guidelines reduced inter-annotator agreement from $\kappa=0.81$ to $0.72$ ($\Delta\kappa \approx 0.09$), despite annotators confirming task equivalence. Similarly, \citep{wiebe2005annotating} documented $\kappa$ reductions of $\approx0.11$.

If LLMaJ is to replace and/or improve the performance of human annotators, a reasonable expectation is that it achieves the same level of robustness as humans. This sets the first baseline ``human robustness'' at $\kappa=0.80 (\Delta 0.07)$ based on the user study described above. However, since LLMs exhibit inherent instability due to floating-point operations, we must establish a stability baseline: what are the reasonable expectations if we do not change the prompts? To establish this stability baseline, we conducted five identical runs over 100 samples per model using fixed prompts. Results showed 89.8\% overlap, with Cohen's $\kappa=0.86$ and Fleiss' $\kappa=0.89$, indicating ``almost perfect agreement'' \citep{kappa}. In contrast, the same runs with $T=0.7$ produce lower agreement (Fleiss' $\kappa=0.39$). This establishes the expected upper bound when the Prompt Lottery is absent and confirms that subsequent instabilities stem from prompt sensitivity rather than stochastic artifacts. The final, most permissive baseline is set at $\kappa > 0.61$, a generally accepted agreement threshold in usability studies per \citep{artstein-poesio-2008-survey, barth2017agreement}.

\section{Experiments} \label{sec:experiments}

 We set up experiments to: (1) quantify judge alignment and the existence of the ``Prompt Lottery'', (2) explore its provenance, (3) its impact on judgment quality, and (4) present mitigation strategies.

\begin{figure*}
    \centering
    \includegraphics[width=0.9\linewidth]{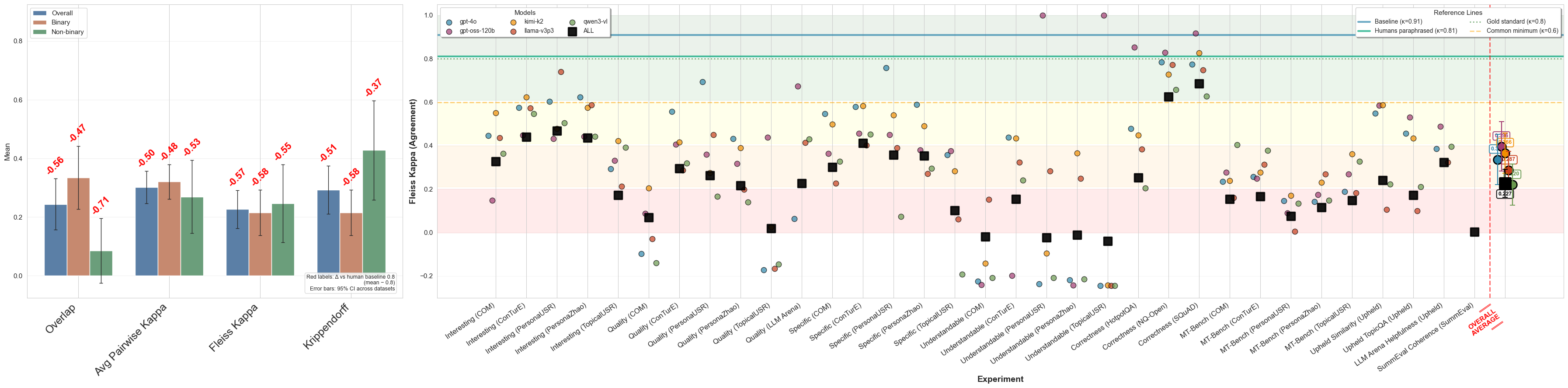}
    \caption{On the left bar-chart: agreement metrics demonstrate low consistency across syntactic variants both across models and prompts. On the right we show Fleiss'~$\kappa$ values and degradation below baselines across judge families for each model. Table~\ref{tab:combined_stability} provides details.}
    \label{fig:agreement_joint}

\end{figure*}
\begin{table*}[h]
\centering
\small
\caption{LLMaJ stability under paraphrastic variation. Fleiss' $\kappa$ values and degradation ($\Delta\kappa = \kappa_{\text{baseline}} - \kappa_{\text{rephrased}}$) across judge families. $\ast$ -- denotes statistical sig. at $0.05$ and $\dagger$ at $0.01$}

\setlength{\tabcolsep}{4pt}
\renewcommand{\arraystretch}{1.1}
 \begin{tabular}{l|cc|cc|cc|cc|cc|cc}
\toprule
\textbf{Judge Family} & 
\multicolumn{2}{c|}{\textbf{GPT-4o}} & 
\multicolumn{2}{c|}{\textbf{Kimi-K2.5}} & 
\multicolumn{2}{c|}{\textbf{Llama 3.3}} & 
\multicolumn{2}{c|}{\textbf{Qwen3-VL}} & 
\multicolumn{2}{c|}{\textbf{GPT-oss-120B}} & 
\multicolumn{2}{c}{\textbf{Model Avg}}  \\
\cmidrule(lr){2-3}\cmidrule(lr){4-5}\cmidrule(lr){6-7}\cmidrule(lr){8-9}\cmidrule(lr){10-11}\cmidrule(lr){12-13}
& $\kappa$ & $\Delta$ & $\kappa$ & $\Delta$ & $\kappa$ & $\Delta$ & $\kappa$ & $\Delta$ & $\kappa$ & $\Delta$ & $\bar{\kappa}$ & $\overline{\Delta}$ \\
\midrule
\textit{Dialogue Quality (Binary)} & & & & & & & & & & & & \\
\quad Interesting & 0.45$\ast$ & 0.43 & 0.55$\ast$ & 0.32 & 0.44$\ast$ & 0.45 & 0.36$\ast$ & 0.54 & 0.15$\ast$ & 0.74$\dagger$ & 0.39$\ast$ & 0.50 \\
\quad Quality & 0.31$\ast$ & 0.58 & 0.41$\ast$ & 0.48 & 0.21$\ast$ & 0.68 & 0.19$\ast$ & 0.71$\dagger$ & 0.16$\ast$ & 0.73$\dagger$ & 0.26$\ast$ & 0.64  \\
\quad Specific & 0.55$\ast$ & 0.33 & 0.50$\ast$ & 0.38 & 0.23$\ast$ & 0.66 & 0.33$\ast$ & 0.57 & 0.36$\ast$ & 0.53 & 0.39$\ast$ & 0.49  \\
\quad Understandable & 0.23$\ast$ & 0.65 & 0.15$\ast$ & 0.75$\dagger$ & 0.35$\ast$ & 0.54 & 0.22$\ast$ & 0.68 & 0.60 & 0.30 & 0.31$\ast$ & 0.58  \\
\midrule
\textit{QA Evaluation (Likert 1-5)} & & & & & & & & & & & & \\
\quad Correctness & 0.47$\ast$ & 0.41 & 0.46$\ast$ & 0.42 & 0.38$\ast$ & 0.51 & 0.20$\ast$ & 0.69 & 0.85 & 0.04 & 0.47$\ast$ & 0.41 \\
\quad Groundedness & 0.55$\ast$ & 0.33 & 0.90 & 0.01 & 0.57$\ast$ & 0.31 & 0.66 & 0.22 & 0.76 & 0.12 & 0.69 & 0.20 \\
\midrule
\multicolumn{13}{l}{\textit{Open-Ended (Likert 1--10)}} \\
\quad MT-Bench & 0.26$*$ & 0.72$\dagger$ & 0.24$*$ & 0.74$\dagger$ & 0.20$*$ & 0.78$\dagger$ & 0.42 & 0.56 & 0.28$*$ & 0.70$\dagger$ & 0.28$*$ & 0.70$\dagger$ \\
\midrule
\multicolumn{13}{l}{\textit{Semantic Alignment (Likert 1--5)}} \\
\quad Content & 0.54$*$ & 0.38 & 0.51$*$ & 0.27 & 0.36$*$ & 0.62 & 0.23$*$ & 0.68$\dagger$ & 0.32$*$ & 0.56$\dagger$ & 0.39$*$ & 0.48 \\
\midrule
\textbf{Model Avg} & \textbf{0.42$\ast$} & \textbf{0.50} & \textbf{0.47$\ast$} & \textbf{0.45$\ast$} & \textbf{0.34$\ast$} & \textbf{0.58} & \textbf{0.34$\ast$} & \textbf{0.58} & \textbf{0.44$\ast$} & \textbf{0.49} & \textbf{0.40$\ast$} & \textbf{0.52} \\
\bottomrule
\end{tabular}
\label{tab:combined_stability}
\end{table*}
\subsection{Agreement Collapses Under Prompt Variation } \label{sec:agreement}

We observe that the agreement degrades from near-perfect (stability and human baseline) to fair ($\kappa$: 0.89 $\rightarrow$ 0.24), 
a drop substantially larger than the 0.09--0.11 degradation we observe in the human studies (similar to the literature and our study). Differences are statistically significant at $p<0.05$ in aggregate, with 75\% of individual dataset-judge configurations showing significant differences. This magnitude of sensitivity points to reliability issues in LLMaJ evaluation setups compared to human annotation protocols. We also observe that on average all but two cases (NQ-Open Correct and SQuAD Correct) fall below a generally accepted agreement level for valid annotation studies ($\kappa<0.61$). This holds for all metrics (~Figure~\ref{fig:agreement_joint}~right), even the highest average agreement, measured by Krippendorff's $\alpha$ for non-binary annotations falls well below both human and theoretical baselines. The agreement is fairly stable, with a 95\% confidence interval half-width of 0.065 on average.

\subsection{ Instability Depends on Task Type and Backbone }

\textbf{The Binary Paradox.} Binary tasks show higher raw overlap (0.32 vs. 0.14 for 
non-binary), suggesting LLMs frequently output identical scores by chance. However, 
chance-corrected reliability (Fleiss' $\kappa=0.214$, $\alpha=0.217$) is lower than for non-binary tasks 
($\kappa=\alpha=0.28$), indicating the output space constraint masks underlying instability 
rather than indicating true robustness.

\textbf{Model-Dependent Sensitivity.} While all models exhibit the Prompt Lottery, magnitude varies. 
Kimi-K2.5 and GPT-4o exhibited greater robustness than Qwen3-VL and Llama 3.3, 
though all models showed substantial degradation from baseline\footnote{Kimi-K2.5: $\kappa=0.47$; GPT-oss-120B: $\kappa=0.44$; GPT-4o:  $\kappa=0.42$; Llama 3.3:  $\kappa=0.34$; Qwen3-VL: $\kappa=0.33$}.
Objective judges (QA Correctness, Groundedness) show higher robustness ($\kappa$=0.47–0.69) compared to subjective 
and open-ended judges ($\kappa$=0.26–0.39), suggesting concrete evaluation criteria are inherently more stable under surface variation. 

\textbf{Newer models.} Reasoning-augmented models introduce internal 
stochasticity orthogonal to temperature control (tree search, best-of-N 
reasoning). This makes temperature-controlled comparison with direct-scoring models problematic. We show that the Prompt Lottery is a universal property of LLMaJ evaluation across both older and frontier models, rather than an artifact of a specific model generation.  We compared GPT-4o and GPT-5.1 scores across identical tasks (5 prompts 
each, 3 datasets, 400 examples). Inter-model agreement (Cohen's $\kappa$) shows a consistent pattern: 
objective tasks ($\kappa$=0.82), dialogue ($\kappa$=0.68), summarization ($\kappa$=0.25). This suggests that both models struggle equally on summarization despite 
architectural differences, and that the bottleneck is task-inherent difficulty that is not solved by more powerful models.

\begin{table}[ht]
  \centering
  \begin{minipage}[t]{0.48\textwidth}
    \centering
    \tiny
            \captionof{table}{Correlation between prompts' $sBERT$ Similarity and Cohen's Kappa.}
\label{tab:sim}
\begin{tabular}{lcccc}
\toprule
Similarity & Pearson $r$ & $p$ & Spearman $\rho$ & $p$ \\
\midrule
Average & $-0.11$ & 0.157 & $-0.15$ & 0.022 \\
Min     & $-0.08$ & 0.25 & $-0.06$ & 0.37 \\
Max     & $0.05$  & 0.44 & $-0.07$ & 0.30 \\
\bottomrule
\end{tabular}
  \end{minipage}%
  \hfill
  \begin{minipage}[t]{0.48\textwidth}
    \centering
    \small
 \captionof{table}{Pearson correlation ($r$) between the default ($p_{1}$) and the top-performing prompts.}

    \begin{tabular}{llll}

    \hline
    \textbf{Dataset} & \textbf{Top-1} & \textbf{Max/$p_{1}$ $r$} & \textbf{Rank} \\ \hline
    Upheld           & ${P_5}$            & 0.44/0.42                 & 4 / 5  \\
    Topical-Chat     & $P_4$            & 0.49/0.47               & 4 / 5\\
    LLM-Arena        & $P_5$            & 0.49/0.49                & 3 / 5\\ 
    \hline
    \end{tabular}
    \label{tbl:individual_correlations}
  \end{minipage}
\end{table}

\subsection{No Prompt Wins Consistently—It's a Lottery }
If LLMaJ relied primarily on semantic content (rather than surface form), high semantic similarity should predict high agreement.
However, we observe sBERT similarities of $0.92-0.95$ yet agreement ($\kappa$) ranges from 0.0 to 0.7, suggesting LLMs are sensitive to non-semantic surface features. The correlation between semantic similarity and scoring stability 
was weak and negative overall ($r=-0.11, p=0.157$ Table~\ref{tab:sim}), indicating that increasing semantic similarity
does not improve judgment consistency. This relationship remained weak and inconsistent even when restricting analysis to the most- and least-agreeable prompt subsets,
 suggesting that semantic similarity is insufficient to explain evaluation stability. Critically, no single prompt variant consistently outperforms others across datasets. The top-performing variant differs per task ($P_5$ on Upheld, $P_4$ on Topical-Chat, $P_5$ on LLM-Arena), and average absolute correlation differences are minimal ($\Delta r = 0.00–0.02$). This absence of a universal ``best'' prompt suggests instability does not stem from systematic prompt content differences (e.g., one being semantically clearer or more complete), but rather from LLM sensitivity to superficial syntactic variation that is fundamentally unpredictable across contexts.

If default prompts are well optimized, they should achieve the highest alignment with human judgments and the observed 
lack of agreement has no bearing; the conclusion would be that LLM practitioners have already optimized the prompts to account 
for variations. If alternative paraphrastic variants consistently outperform, this would suggest sensitivity to prompt formulation, and 
that there is an element of chance associated with single prompt LLMaJ setups. 
To measure this we followed an established protocol in \citep{g-eval} and calculated Pearson correlation ($r$) between human and paraphrased-prompts judgments.
For datasets for which human annotations are publicly accessible (Upheld, Topical-Chat, LLM-Arena) the results 
in Table~\ref{tbl:individual_correlations} show that the latter is the case. Paraphrastic variations consistently match or outperform $P_1$ (default) prompts. 
While the default prompt ranked $3rd -4th$ across datasets, absolute differences in correlation were small ($\Delta r$ = 0.00-0.02), 
and only 2 of 3 comparisons reached significance ($p<0.05$, uncorrected). Bootstrap rank CIs for human alignment overlap fully across all five prompts, and the default prompt is ranked first in only 5--24\% of resamples depending on the human aggregation.
The results in Section~\ref{sec:agreement} suggest that this alignment may be an artifact of aggregation. 
The fact that divergent individual judgments can produce similar aggregate correlations may obscure 
instability in the evaluation process.  This may reconcile our findings with earlier reports of instruction robustness in LLM evaluation~\citep{chiang-lee-2023-large}.

\subsection{Do Prompts Reverse Model Rankings?}

While the Prompt Lottery substantially alters absolute scores, agreement with humans seems to be largely unaffected, but a critical question 
is whether it also reverses \emph{relative model rankings}. This distinction matters: 
score variance may reduce confidence in conclusions, but ranking reversals directly 
contradict them. To test whether this instability affects 
comparative evaluations, we evaluated four inference models 
(GPT-4o, gpt-oss-120B, Kimi-K2.5, and Qwen3p6-Plus) under five paraphrastic 
prompt variants using a fixed backbone model.
All ranking stability statistics were computed over 100 evaluation 
samples per task and averaged across three independent folds to reduce sampling variance and isolate 
prompt-induced ranking instability. 

Ranking reversals are strongly task-dependent. Objective evaluation 
(SQuAD v2 correctness) exhibits severe instability ($\tau = 0.18$;  model rank positions change in 43.3\% of prompt comparisons). For this task GPT-4o and Qwen3p6-Plus models exchange top rankings across prompt variants, with GPT-4o ranging from 1st to 3rd 
and Qwen3p6-Plus from 1st to 4th. In contrast, subjective dialogue evaluation 
(MTBench) remains perfectly stable (0\% reversals). Intermediate behavior appears in 
SummEval and persona-based dialogue evaluation, where rankings remain partially 
stable but top-performing systems still change depending on prompt wording.  The identity of the ``best'' model thus depends 
substantially on prompt formulation.

We visualize how judgments vary when, for the same input $I$, either the back-model or prompt changes.

\begin{itemize}
    \item $\Delta$ $XX\%$: fraction where row scores higher than col;
    \item $\equiv$ $XX\%$: fraction of exact matches;
    \item $\nabla$ $XX\%$: fraction where row scores lower than col;
\end{itemize}

\begin{figure*}[h]
    \centering
    \begin{minipage}[t]{0.48\textwidth}
        \centering
        \includegraphics[width=\linewidth]{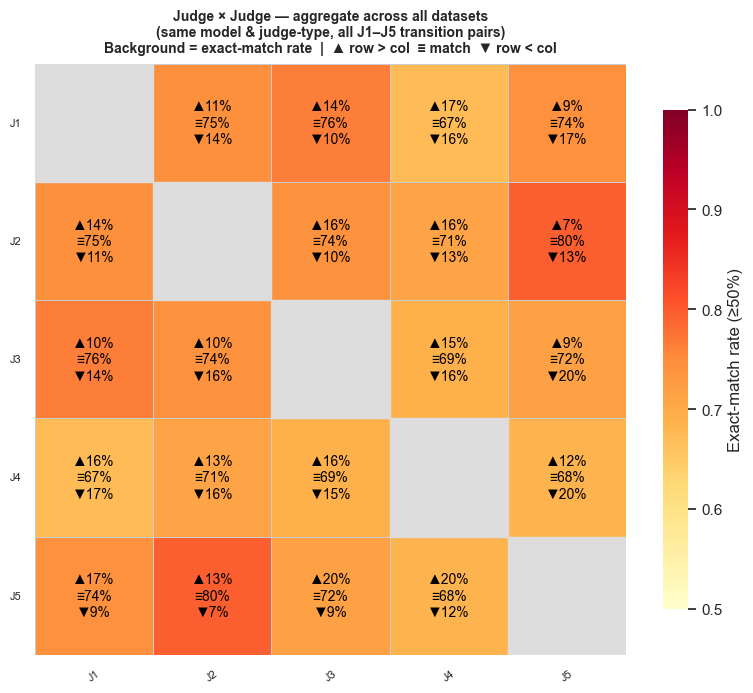}
        \caption{Prompt-vs-Prompt: Judge-vs-Judge aggregated across all experimental setups.}
        \label{fig:heat_judge}
    \end{minipage}%
    \hfill
    \begin{minipage}[t]{0.48\textwidth}
        \centering
        \includegraphics[width=\linewidth]{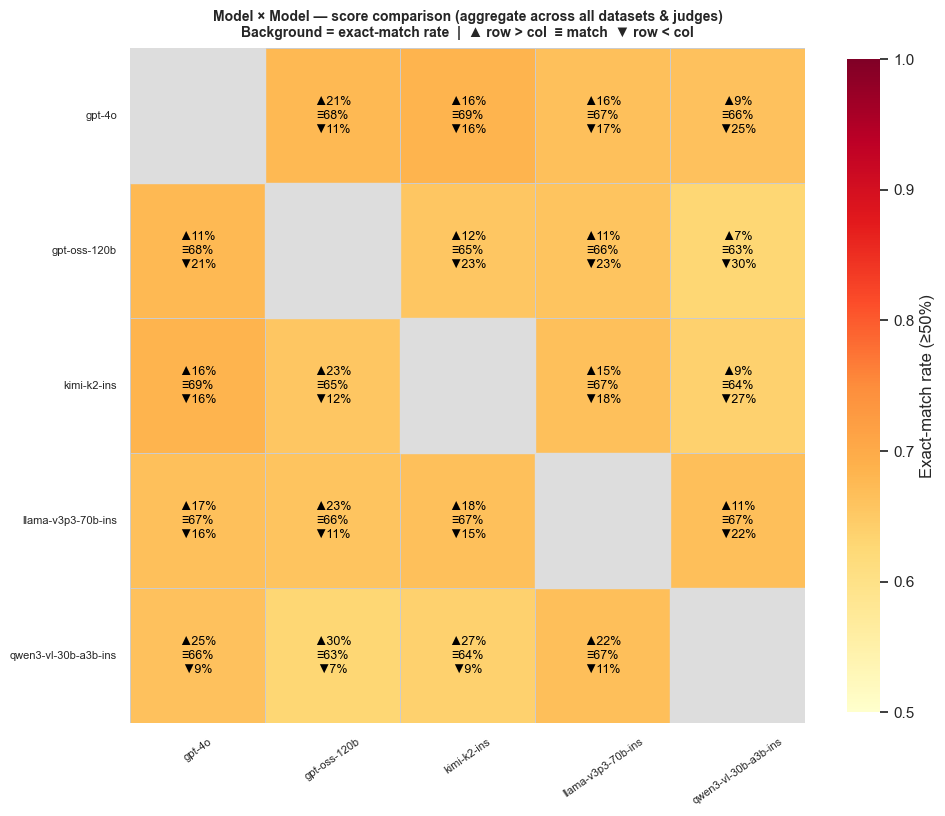}
        \caption{Model-vs-Model: Compares models while holding the judge type and number fixed, aggregated across all datasets.}
        \label{fig:model_vs_model}
    \end{minipage}
\end{figure*}

\begin{table}[h]
\small
\centering
\captionof{table}{Ranking stability across paraphrastic prompts. Lower Kendall's $\tau$ and higher absolute reversal rates indicate stronger Prompt Lottery effects.}

\begin{tabular}{lccc}

\toprule
\textbf{Task} & \textbf{Mean $\tau$} & \textbf{Top-1 Flips} & \textbf{Flip \%} \\
\midrule
MTBench (DailyDial.) & 1.00 & 0/4 & 0.0\% \\
MTBench (Persona) & 0.80 & 2/4 & 10.0\% \\
QA Correct. (SQuAD v2) & 0.18 & 3/4 & 43.3\% \\
Quality (SummEval) & 0.64 & 2/4 & 40.0\% \\
G-Eval (Topical-Chat) & 0.83 & 1/4 & 10.0\% \\
\bottomrule
\end{tabular}
\label{tab:ranking_stability}
\end{table}

\section{Mitigating via Aggregation} \label{sec:mitigation}

\begin{figure}[h]
  \centering
  \begin{minipage}{0.27\textwidth}
    \centering
    \includegraphics[width=\linewidth]{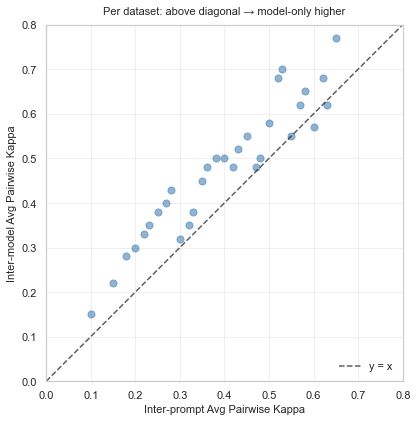}
    \caption{Model vs prompt.}
    \label{fig:model_vs_prompt_sub}
  \end{minipage}%
  \hfill
  \begin{minipage}{0.27\textwidth}
    \centering
    \includegraphics[width=\linewidth]{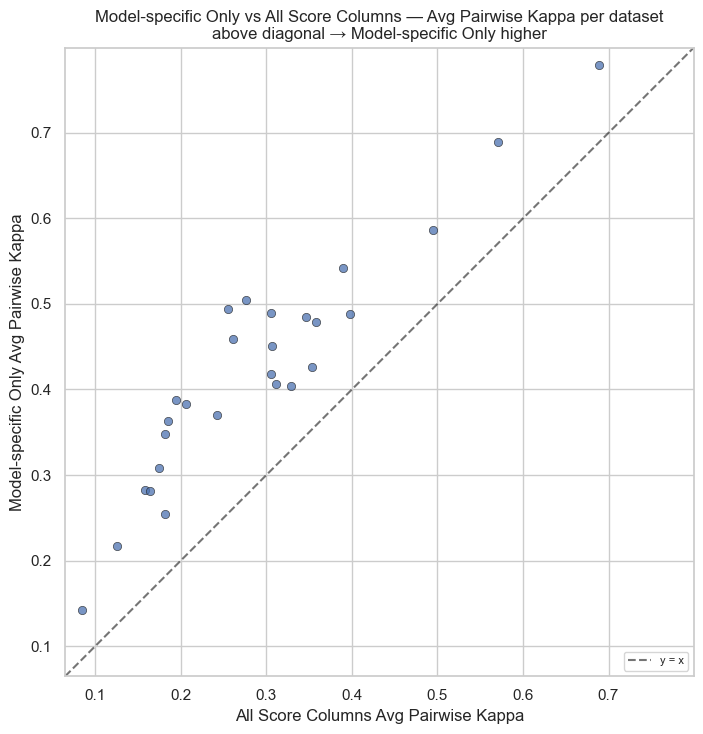}
    \caption{Model vs hybrid.}
    \label{fig:model_vs_hybrid}
  \end{minipage}%
  \hfill
  \begin{minipage}{0.44\textwidth}
    \centering
    \small
            \captionof{table}{Pearson correlation ($r$) between human scores and LLMaJ aggregation strategies ($p<0.05$).}

    \begin{tabular}{@{}lcccc@{}}

    \toprule
    \textbf{Dataset} & \textbf{Maj.} & \textbf{Avg.} &  \textbf{Plur.} & \textbf{PA.}\\ 
    \midrule
    \scriptsize
    Upheld       &0.47&0.54&0.63&0.61\\
    \tiny $\rightarrow$95\%~CI&\scriptsize{[.42,.57]}&\scriptsize[.48,.62]&\scriptsize[.52,.69]&\scriptsize[.48,.62]\\
    \scriptsize Topical-Chat  &0.44&0.55&0.62&0.51\\
    \tiny $\rightarrow$95\%~CI &\scriptsize [.36,.48] & \scriptsize[.50,.59] & \scriptsize[.57,.69] &\scriptsize[.43,.53] \\
    \scriptsize LLM-Arena     &0.44&0.54&0.55&0.52\\
    \tiny $\rightarrow$95\%~CI & \scriptsize[.32,.48] & \scriptsize[.47,.57] & \scriptsize[.47,.59] &\scriptsize[.45,.53] \\
    \bottomrule
    \end{tabular}
    \label{tbl:aggregation_results}
  \end{minipage}
\end{figure}

Prior work has implicitly addressed judge instability via LLMaJ aggregation \citep{zhao2025voices, mizrahi2024stateartmultipromptllm}.
Motivated by this, we evaluate which aggregation strategy across paraphrastic prompt variants best aligns with human judgments using the three datasets with public human annotations. We compare \textbf{Average}, \textbf{Majority} (mode), \textbf{Plurality} (requiring $>50\%$ consensus), and \textbf{Plurality+Avg. (PA)} (using average when plurality is not reached)
using Pearson correlation ($r$) with human scores.
Table~\ref{tbl:aggregation_results} shows Plurality consistently achieves the strongest human alignment, improving over both averaging and majority voting. 
This points to a possible strategy, often used in human studies, where data points are discarded if they do not reach a certain threshold of agreement.

\subsection{Which Source of Aggregation Matters?}

Having established that aggregation improves alignment with human judgments, we next ask which source of diversity contributes most to these gains. Prior work has largely emphasized model diversity (e.g., through ensembles or debate-style systems), while our findings suggest that prompt variation itself may provide a complementary source of robustness. 
To examine this, we compare three aggregation strategies: (1) \textbf{Inter-Model} (varying models, fixed prompts), (2) \textbf{Inter-Prompt} (varying prompts, fixed models), and (3) \textbf{Hybrid} (varying both models and prompts). Figures~\ref{fig:model_vs_prompt_sub} and~\ref{fig:model_vs_hybrid} show that inter-prompt aggregation achieves stronger reliability compared to the other two, suggesting that prompt variation may contribute more strongly to robustness than model diversity alone. Our aggregation strategies assume the five paraphrased-prompt judgments are conditionally independent given the true label. Because all five variants share the same backbone model and task, this assumption may not hold in practice; dependence-aware aggregation methods \citep{balasubramanian2026dependence} show that independence-assuming rules can be suboptimal. 

\subsection{Selective Aggregation: Task-Dependent Cost Savings} \label{sec:cost}

We analyze when the default prompt aligns with broader consensus. Binary tasks show high stability and match plurality consensus in 94.7\% of cases, making multi-prompt aggregation marginally useful due to limited output space. Non-binary tasks exhibit the opposite, and across all configurations, strict plurality consensus ($>50\%$ agreement) emerges in only 58.73\% of cases, indicating that many examples fail to converge even after evaluating five prompt variants. Moreover, the default prompt aligns with the plurality decision in only 44.0\% of cases, implying that more than half of single-prompt evaluations disagree with the consensus whenever prompt variation is considered. These results suggest that single-prompt evaluation can diverge substantially from aggregated judgments. Consensus statistics in non-binary tasks suggest many examples either converge early or fail to converge entirely: 37.68\% reach plurality by the third judge, 21.05\% require all five, and 41.27\% never converge. This distribution suggests that substantial inference cost is spent on examples that either converge early or fail to converge entirely.
We formulate selective aggregation as a binary classification task: given a prompt 
and an input evaluated, predict whether the initial judgment will match the plurality consensus 
(no aggregation needed, label=0) or whether additional prompt variants should be 
evaluated (label=1). Labels derive from observed plurality outcomes in our experiments.

We evaluated multiple classifiers for selective aggregation, including transformer encoders, classical models, and zero-/few-shot LLM baselines. Architecture selection (Table~\ref{tab:models}) was performed on a smaller default split, and the selected architecture was then retrained on $N = 369{,}625$ samples with hyper-parameters selected via Optuna \citep{optuna_2019} over 35 trials, and evaluated on a held-out test set of $n = 73{,}925$ samples. Logistic Regression and SVM serve as baselines to assess whether prompt stability can be captured with lightweight decision boundaries, while zero-/few-shot LLM variants represent a potentially more expressive alternative that may capture semantic signals at higher computational cost. While \textsc{RoBERTa}~\citep{liu2019roberta} achieved the highest balanced accuracy, we selected \textsc{DeBERTa-V3} \citep{he2021debertav3} because our objective is to maximize true negative rate (TNR), rather than overall performance. The final classifier achieves TPR 85.06\% and TNR 69.42\%, deliberately optimizing in this way as false positives incur only inference cost, while false negatives corrupt evaluation quality. To assess generalization, we evaluate on a held-out dataset with Prometheus judges~\citep{kim2023prometheus} whose prompts were absent from training. Performance suggests the classifier captures signals that generalize beyond the training prompts and datasets with TNR stable at 66.22\%. More details on model training are reported in Appendix~\ref{sec:model-params}.

\begin{table}[t]
  \centering
  \begin{minipage}[t]{0.45\textwidth}
    \centering
    \small
    \setlength{\tabcolsep}{4pt}
     
         \captionof{table}{Model ablation results on a smaller dataset with prompts not in the training data.}

    \begin{tabular}{@{}lcc@{}}
    \toprule
    Model & Bal. acc & TNR \\
    \midrule
    \textsc{DeBERTa} & 0.836 & 0.854 \\
    \textsc{RoBERTa} & 0.914 & 0.845 \\
    Logistic & 0.862 & 0.751 \\
    SVM & 0.854 & 0.742 \\
    Llama zero-shot & 0.489 & 0.042 \\
    5-shot & 0.451 & 0.322 \\
    \bottomrule
    \end{tabular}
    \label{tab:models}
  \end{minipage}%
  \hfill
  \begin{minipage}[t]{0.50\textwidth}
    \centering
    \small
    \captionof{table}{Cost-benefit strategy comparisons. Worst-case scores are those with no plurality consensus. $C$ stands for compute power. $Plu.~\%$ is the share of data points with a plurality decision.}
    \begin{tabular}{lrrr}
    \toprule
    \textbf{Strategy} & \textbf{$C$} & \textbf{$Plu.~\%$} & \textbf{Worst-case} \\
    \midrule
    Single LLMaJ      & 1$\times$  & 44\% & -- \\
    Task heuristic & $\sim$4.5$\times$ & 74\% & 69\% \\
    \textsc{DeBERTa}-gate & $\sim$2$\times$ & 86\% & 65\% \\
    Full aggregation  & 5$\times$  & 100\% & 79\% \\
    \bottomrule
    \end{tabular}
   
    \label{tab:cost}
  \end{minipage}
\end{table}

In our selective aggregation framework, cost savings occur when stable instances are correctly identified as negatives, allowing aggregation to be skipped. False positives unnecessarily trigger additional LLM calls and reduce efficiency gains. We compare against a \textbf{task-type heuristic}: always aggregate non-binary 
tasks (44\% instability), never aggregate binary tasks (94.7\% stability). While 
this strategy is cheaper than full aggregation for stable tasks, it wastes compute 
on non-binary instances and misses unstable binary cases. Our \textsc{DeBERTa} 
approach provides instance-level granularity that the heuristic lacks. As shown in Table~\ref{tab:cost}, \textsc{DeBERTa}-gating reduces LLM calls by approximately 60\% compared to naive aggregation ($2 \xrightarrow{} 5\times$ compute) while recovering 86\% of full aggregation decisions. Performance declines to 65\% only in difficult no-consensus cases, representing the maximum observed degradation; typical performance is substantially higher. Full aggregation achieves 79\% even in worst-case settings because all 
prompt variants are evaluated directly. 

\subsubsection{Derivation of LLM Call Reduction Under Selective Aggregation}
\label{appendix:cost-derivation}

\textbf{Step 1: Expected calls under full aggregation.} From the empirical distribution of convergence times reported in
Section~\ref{sec:cost}, non-binary tasks require the following number of judge
calls to determine the plurality outcome. Plurality is reached at three judge calls for 37.68\% of cases,
at five calls for 21.05\%, and 41.27\% of cases require more than five calls to reach plurality.

\begin{equation}
\bar{t}_{\text{full}}
  > 0.3768 \times 3 + 0.2105 \times 5 + 0.4127 \times 5
  = 4.247
\label{eq:t-full}
\end{equation}

\textbf{Step 2: Expected calls under selective aggregation.} The classifier achieves a TNR of 69.42\%. The
remaining cases go through the aggregation pipeline.

\begin{equation}
\begin{split}
\bar{t}_{\text{gated}}
  = \underbrace{\mathrm{TNR} \times 1}_{\text{single-prompt-cases}}
  + \underbrace{(1 - \mathrm{TNR}) \times \bar{t}_{\text{full}}}_{\text{aggregated-cases}} 
  = 0.6942 + 0.3058 \times 4.247
  \approx 1.99
\end{split}
\label{eq:t-gated}
\end{equation}

\textbf{Step 3: Call reduction.} Relative to full aggregation ($\bar{t}_{\text{full}} = 4.247$):

\begin{equation}
\Delta_{\text{vs.\ full}}
  = 1 - \frac{\bar{t}_{\text{gated}}}{\bar{t}_{\text{full}}} > 1 - \frac{1.99}{4.247}
  \approx 53\%
\label{eq:reduction-full}
\end{equation}

Relative to the na\"{i}ve five-prompt baseline
($\bar{t}_{\text{naive}} = 5.0$) the $\Delta_{\text{vs.\ naive}}$ would be~$\approx 60\%$.

\section{Limitations and Final Remarks}

We find that LLMaJ outputs are sensitive to paraphrastic prompt variation, 
affecting absolute scores and relative model rankings. We further 
show that this instability can reverse relative rankings of evaluated models, 
particularly in objective QA settings, demonstrating that prompt choice may alter 
both absolute and comparative evaluation conclusions. We show that plurality aggregation improves agreement with humans compared to 
single-prompt evaluation. By computing consensus across semantically equivalent 
prompts, the plurality metric filters pragmatic noise from sub-optimal variations. 
This significantly improves alignment with human judgments, confirming that LLMaJ 
reliability is maximized when multiple equivalent prompts converge. While 
aggregation improves robustness, it increases computational cost. Our selective 
aggregation approach makes this practical by predicting which judgments require 
additional evaluations. Using a \textsc{DeBERTa} classifier to identify stable instances, we reduce LLM calls from 5 to 1.99 
transforming multi-prompt evaluation into a viable, cost-effective default strategy.

\paragraph{Scope and Limitations.}
Our findings are limited to paraphrastic variations that preserve semantic content; 
we do not test deliberate semantic modifications (e.g., changing evaluation criteria), 
which are addressed elsewhere. Semantic equivalence is 
validated via sBERT similarity (0.93) and human task equivalence (\(\kappa=0.78\)) 
rather than formal logical entailment. Additionally, our main experiments use 
five model families, with 
limited coverage of reasoning-augmented models; a preliminary comparison of GPT-4o 
and GPT-5.1 (400 examples, 3 datasets) suggests Prompt Lottery persists across 
architectures. However, the magnitude of 
degradation (\(\kappa: 0.89 \to 0.24\)) substantially exceeds human variability 
(\(\Delta\kappa = 0.09\)), and findings persist across all experimental conditions, 
suggesting Prompt Lottery effects are robust despite these constraints. Our mitigation strategy is evaluated against single-prompt (lower bound) and 
full aggregation (upper bound) baselines. We do not compare against budget-matched alternatives (e.g., single-prompt self-consistency sampling) or orthogonal robustness techniques (ensembles, debate, instruction tuning), so the gains in Section~\ref{sec:mitigation} may partly reflect repeated judging rather than prompt diversity per se; this analysis is reserved for future work.

From a theoretical standpoint, these results reveal a departure from 
\textit{semantic invariance}, a foundational principle from formal semantics predicting that rational judges should produce 
equivalent outputs for semantically equivalent inputs. This pragmatic behavior, while potentially adaptive 
for dialogue understanding, creates brittleness in evaluation. Taken together, our findings support the central methodological 
message of this paper: evaluation should move beyond the single-prompt paradigm and 
toward systematic multi-prompt assessment. Robustness to prompt formulation is not 
an optional refinement, but a prerequisite for trustworthy LLMaJ evaluation and 
reliable comparative model ranking. For practitioners, we summarize these findings 
in Table~\ref{tab:takeaways}; for researchers, it means treating prompt sensitivity 
as a first-class concern in evaluation methodology, not an afterthought.

\begin{table}[h]
\small
\centering
\caption{Takeaways for robust LLMaJ evaluation.}
\begin{tabular}{lc}
\toprule
\textbf{Factor} & \textbf{Key Finding} \\
\midrule
Backbone model choice & Largest source of pairwise agreement variance \\
Prompt rephrasing & Significant variance despite semantic equivalence \\
Binary evaluation tasks & High stability, skip aggregation \\
Non-binary evaluation tasks & High instability, aggregate \\
Plurality aggregation & Strong robustness improvement \\
Selective aggregation & Major cost reduction (\(\sim 60\%\)) \\
\bottomrule
\end{tabular}
\label{tab:takeaways}
\end{table}

\bibliographystyle{unsrt}
\bibliography{custom.bib}

\clearpage
\appendix
\section{Mapping of Judge Families to Dataset and Tasks} \label{appendix:experiments}

\begin{table*}[h]
\centering
\small
\caption{%
  Overview of the 33 experimental configurations used in this study.
  Each configuration pairs a dataset with a judge family and is evaluated
  with $5 \times 5 = 25$ distinct \textsc{LLMaJ} triples
  (5 back-models $\times$ 5 prompt variants).
  \emph{Binary?}\ indicates whether the judge produces a binary (Yes/No)
  rather than a Likert-scale score.
  Judge-family abbreviations---\textbf{DQ}: Dialogue Quality;
  \textbf{GQ}: General Quality (MT-Bench style);
  \textbf{QC}: QA Correctness;
  \textbf{QG}: QA Groundedness;
  \textbf{SA}: Semantic Alignment;
}
\setlength{\tabcolsep}{5pt}
\renewcommand{\arraystretch}{1.15}
\begin{tabular}{llcc r}
\toprule
\textbf{Dataset} & \textbf{Judge family (dimension)} & \textbf{Binary?} & \textbf{Configs} & \textbf{$n$} \\
\midrule
 
\multirow{5}{*}{PersonaChat}
  & DQ — Interesting      & \checkmark & 1 & 149 \\
  & DQ — Quality          & \checkmark & 1 & 149 \\
  & DQ — Specific         & \checkmark & 1 & 149 \\
  & DQ — Understandable   & \checkmark & 1 & 149 \\
  & GQ — MT-Bench         &            & 1 & 149 \\
\midrule
 
\multirow{5}{*}{PersonaUSR}
  & DQ — Interesting      & \checkmark & 1 & 60 \\
  & DQ — Quality          & \checkmark & 1 & 60 \\
  & DQ — Specific         & \checkmark & 1 & 60 \\
  & DQ — Understandable   & \checkmark & 1 & 60 \\
  & GQ — MT-Bench         &            & 1 & 60 \\
\midrule
 
\multirow{5}{*}{Topical-Chat}
  & DQ — Interesting      & \checkmark & 1 & 60 \\
  & DQ — Quality          & \checkmark & 1 & 60 \\
  & DQ — Specific         & \checkmark & 1 & 60 \\
  & DQ — Understandable   & \checkmark & 1 & 60 \\
  & GQ — MT-Bench         &            & 1 & 60 \\
\midrule
 
\multirow{5}{*}{ConTurE}
  & DQ — Interesting      & \checkmark & 1 & 1{,}013 \\
  & DQ — Quality          & \checkmark & 1 & 1{,}013 \\
  & DQ — Specific         & \checkmark & 1 & 1{,}013 \\
  & DQ — Understandable   & \checkmark & 1 & 1{,}013 \\
  & GQ — MT-Bench         &            & 1 & 1{,}013 \\
\midrule
 
DailyDialog
  & GQ — MT-Bench         &            & 1 & 100 \\
\midrule
 
\multirow{2}{*}{HotpotQA}
  & QC — Correctness      &            & 1 & 1{,}000 \\
  & QG — Groundedness     &            & 1 & 1{,}000 \\
\midrule
 
\multirow{2}{*}{SQuAD v2}
  & QC — Correctness      &            & 1 & 1{,}000 \\
  & QG — Groundedness     &            & 1 & 1{,}000 \\
\midrule
 
\multirow{2}{*}{NQ-Open}
  & QC — Correctness      &            & 1 & 1{,}000 \\
  & QG — Groundedness     &            & 1 & 1{,}000 \\
\midrule
 
SummEval
  & Coherence, Consistency, Fluency &  & 3 & 160 \\
\midrule

\multirow{1}{*}{Topical QA}
  & SA — Content        &            & 1 & 1{,}000 \\

 \midrule
\multirow{1}{*}{LLM-arena}
   & SA — Content        &            & 1 & 1{,}000 \\

  \midrule
  \multirow{1}{*}{Upheld}
   & SA — Content        &            & 1 & 1{,}000 \\

\midrule
 
\addlinespace[2pt]
\textbf{12 datasets} & & & \textbf{33} & \\
\bottomrule
\end{tabular}

\label{tab:experimental-configurations}
\end{table*}

\section{Dimensions of Paraphrastic Variation} \label{appendix:prompt_diminsions}

After examining paraphrastic variations post hoc across default prompts from all judge families, we distilled six representative dimensions. The six paraphrastic dimensions (Appendix B) were identified by post hoc inspection of the generated variants rather than imposed as generation constraints. These are shown in Table~\ref{tab:paraphrastic_dimensions}.

\begin{table*}[h]
\centering
\caption{Qualitative Analysis of Judge Prompts} \label{appendix:prompt_dim}

\small 
\begin{tabular}{l p{4.5cm} p{4.5cm}}
\toprule
\textbf{Dimension} & \textbf{Description} & \textbf{Examples} \\
\midrule
1. Role Framing & How the judge's identity/authority is established. & ``You are an expert evaluator'' vs. ``Act as a judge'' vs. ``You evaluate'' \\
\addlinespace
2. Instruction Syntax & Imperative vs. declarative vs. modal phrasing. & ``Compare...'' vs. ``You should compare...'' vs. ``Your task is to compare...'' \\
\addlinespace
3. Criteria Presentation & How evaluation standards are formatted. & Bullet list vs. numbered list vs. prose vs. scale labels \\
\addlinespace
4. Output Format Specification & Strictness of format requirements. & ``strictly following this format'' vs. ``using exactly:'' vs. ``in the format:'' \\
\addlinespace
5. Task Contextualization & How much context/setup precedes the core task. & Verbose explanation vs. direct instruction \\
\addlinespace
6. Linguistic Register & Formality level and vocabulary choices. & ``assess'' vs. ``evaluate'' vs. ``rate'' vs. ``determine'' / ``brief'' vs. ``short'' vs. ``concise'' \\
\bottomrule
\end{tabular}
\label{tab:paraphrastic_dimensions}

\end{table*}

Each judge prompt was coded on the six dimensions using GPT-5.0 and matched to mean pairwise Cohen's $\kappa$ among models scoring the same items with that prompt (50 prompts, 46 experiments, without $G_{15}$). Feature labels were permuted \emph{within} each experiment--family block (10{,}000 draws per contrast), so the task, items, and model set stayed fixed; associations were averaged with equal weight across families and tested with Holm correction over the seven primary contrasts. Table~\ref{tab:prompt_stability} summarizes the results after residualizing out whether the prompt included a reference answer (which is not one of the six dimensions). Role, syntax, criteria layout, length, verb choice, and brevity all matched the permutation null; their eta-squared values (0.26--0.48) were comparable to the chance baseline for a random grouping of five paraphrases (null means 0.26--0.48). Output-format strictness was the only univariate slope that rejected the null ($+0.076$ $\kappa$ per step on a four-level code from loose to ``strictly following''; Holm $p = 0.0007$), but this did not survive as a format effect: the MT-Bench family written to span those phrases had slope $+0.005$, and holding the other coded features fixed reduced the slope to $+0.031$ ($p = 0.17$). Two listed framings (``You evaluate'' and ``you should compare'') do not occur in this prompt set. Overall, none of the six dimensions identifies wording that consistently raises cross-model score agreement.

\begin{table*}[h]
\centering
\small
\caption{Within-family permutation tests of paraphrastic prompt dimensions against cross-model Cohen's $\kappa$. Permutation $p$-values are two-sided for slopes and one-sided for eta-squared; Holm adjustment is over the seven contrasts.}
\begin{tabular}{l l r r r r}
\toprule
\textbf{Dimension} & \textbf{Estimand} & \textbf{Value} & \textbf{Perm.\ $p$} & \textbf{Holm $p$} & \textbf{Families} \\
\midrule
Role framing & Mean within-family $\eta^2$ & 0.34 & 0.34 & 1.00 & 8 \\
Instruction syntax & Mean within-family $\eta^2$ & 0.32 & 0.78 & 1.00 & 8 \\
Criteria presentation & Mean within-family $\eta^2$ & 0.26 & 0.45 & 1.00 & 7 \\
Output-format strictness & Slope ($\kappa$ per step) & $+0.076$ & $<0.001$ & 0.0007 & 4 \\
Task contextualization (length) & Slope ($\kappa$ per 100 words) & $-0.044$ & 0.63 & 1.00 & 10 \\
Linguistic register (verb) & Mean within-family $\eta^2$ & 0.48 & 0.51 & 1.00 & 10 \\
Linguistic register (brevity) & Mean within-family $\eta^2$ & 0.42 & 0.77 & 1.00 & 2 \\
\bottomrule
\end{tabular}

\label{tab:prompt_stability}
\end{table*}
\section{Semantic Invariance in Language Understanding: Theoretical Foundations}\label{app:semantic_invariance}

A foundational principle in formal semantics is \textbf{semantic invariance}: two linguistic expressions with identical truth conditions should produce equivalent outputs in rational language-processing systems. This principle—rooted in ~\citep{frege1884}, ~\citep{church1951}, and \citep{montague1974}—motivates the expectation that LLM-as-a-Judge systems should be robust to paraphrastic (semantically equivalent) prompt variations. We briefly develop this principle and explain why the Prompt Lottery phenomenon represents a departure from these theoretical expectations.

\subsection{Theoretical Basis for Semantic Invariance}\label{subsec:theory_basis}

Three theoretical commitments support the expectation of semantic invariance in language understanding systems:

\noindent\textbf{Compositionality.} Frege's principle of compositionality~\citep{frege1884,partee1984} states that the meaning of a complex expression depends on the meanings of its parts and the rules combining them. Under this view, two prompts with identical component meanings and combination rules should yield identical results. For example, a prompt asking ``Is the response understandable?'' and one asking ``Is the response clear and easy to comprehend?'' share the same propositional content, differing only in salience and cognitive framing.

\noindent\textbf{Representation Independence.} Church's thesis in semantics~\cite{church1951} asserts that semantic content is independent of representation. A rational agent should produce equivalent outputs given two different encodings of the same semantic content. This principle motivated Montague's formalization of natural language semantics~\cite{montague1974}, where truth conditions are computed independently of surface form.

\noindent\textbf{Rational Agent Theory.} \citep{dennett1987} and \citep{newell1976} establish that a rational agent's outputs should depend on the content of its beliefs and goals, rather than their specific representation. Consequently, representation-invariant systems exhibit higher rationality and generalization.

\subsection{Why Semantic Invariance Might Fail: Pragmatics}\label{subsec:pragmatics}

However, pragmatics—the study of meaning in context—provides complementary explanations for why semantic invariance should not be expected in real-world language processors:

\noindent\textbf{Gricean Implicature.} Grice's Cooperative Principle~\cite{grice1975} characterizes communication as the cooperative coordination of speaker intent. Deviations from Gricean maxims (quantity, quality, manner, and relevance) generate implicatures: listener inferences about unspoken speaker intent. A prompt that elaborates ``clear \textit{and easy} to comprehend'' implicates that comprehensibility is non-obvious, which subtly raises evaluation thresholds. LLMs sensitive to implicature will diverge on such variants despite identical semantic content.

\noindent\textbf{Cognitive Salience.} Relevance Theory~\cite{sperber1986} posits that communication involves trade-offs between cognitive effort and inferential payoff, meaning explicit information is processed more readily than implicit information. A prompt stating ``1 = none of the reference information is conveyed; 5 = semantically equivalent'' makes scale meaning more cognitively available than a prompt stating ``score on a scale of 1 to 5'' alone. LLMs trained on natural language learn that salient, explicit information is more reliable, inducing representation dependence.

\noindent\textbf{Register and Authority Framing.} Register—the style of language appropriate to a context~\cite{hymes1972}—conveys social and communicative intent. Prompts like ``You are an expert evaluator...'', ``Act as a judge...'', or ``You evaluate...'' specify identical tasks but differ in formality, deference, and authority claims. LLMs that model speaker identity and context (necessary for dialogue systems) extract distinct signals from these register cues, inducing pragmatic sensitivity.

\subsection{Reconciling Theory and Empirics}\label{subsec:reconcile}

Our findings reveal that LLMs violate semantic invariance: agreement degrades from $\kappa = 0.89$ (fixed prompt) to $\kappa = 0.24$ (paraphrastic variants) despite a high semantic similarity ($> 0.93$ via sBERT). This violation is consistent with two theoretical frameworks:

\begin{enumerate}
\item \textbf{LLMs are pragmatically sensitive.} They are not purely semantic processors; instead, they extract and weight pragmatic cues (salience, register, and implicature) alongside propositional content. While this is \textit{adaptive} in natural language understanding, it induces \textit{evaluation brittleness}.
\item \textbf{Compositionality may be imperfect.} Our data show that objective tasks (with clear truth conditions) are more invariant ($\kappa = 0.47$--$0.69$) than subjective tasks ($\kappa = 0.26$--$0.28$). This pattern aligns with compositionality theory: systems that learn compositional structure should show stronger invariance when ground truth is unambiguous, whereas ambiguity amplifies pragmatic inference.
\end{enumerate}

Our multi-prompt aggregation strategy (Section~\ref{sec:mitigation}) operationalizes these insights: by averaging across pragmatic variants, we filter out pragmatic noise while preserving the semantic signal. This approach assumes LLMs \textit{are} pragmatically sensitive and designs around that fact, rather than expecting unrealistic invariance.

In summary, semantic invariance is a principled theoretical expectation grounded in formal semantics and rational agent theory. The Prompt Lottery violates this principle because LLMs are fundamentally pragmatically sensitive systems. Understanding this gap is essential both for building trustworthy evaluation pipelines and for broader questions regarding LLM rationality and language understanding.

\section{Agreement Across Models and Prompts}

These results suggest that LLM evaluators exhibit substantial sensitivity to paraphrastic reformulations that humans often perceive as semantically equivalent. If Prompt Lottery effects were absent, we would expect a relatively uniform distribution of high $\kappa$ values across the x-axis. Instead, substantial variance in $\kappa$ despite nearly identical similarity levels suggests the presence of Prompt Lottery effects, where prompt variants intended to preserve semantic equivalence can produce meaningfully different evaluation outcomes.

\begin{figure*}[h]
    \centering
    \begin{minipage}[t]{0.48\textwidth}
        \centering
        \includegraphics[width=\linewidth]{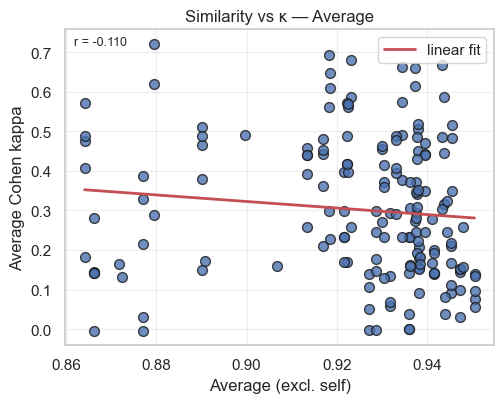}
        \caption{All cases: Comparison across average judge agreement.}
        \label{fig:bert_kappa_avg}
    \end{minipage}%
    \hfill
    \begin{minipage}[t]{0.48\textwidth}
        \centering
        \includegraphics[width=\linewidth]{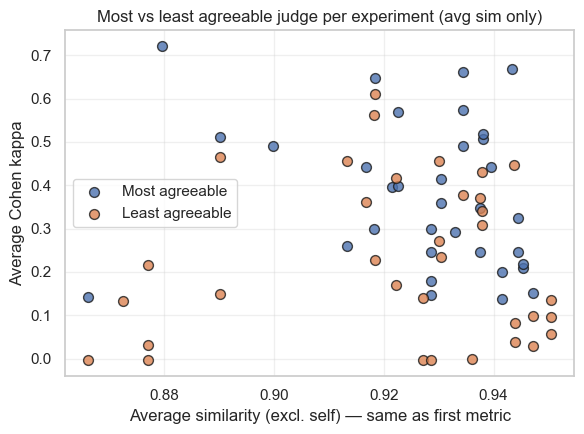}
        \caption{Extreme Cases: Comparison across extreme judge agreement.}
        \label{fig:kappa_vs_bert_x}
    \end{minipage}
\end{figure*}
\include{apendix/appedinx_heatmap.tex}

\section{Rank reversal experiments} \label{appendix:ranks}

\begin{figure*}[h]
    \centering
    \begin{minipage}[t]{0.48\textwidth}
        \centering
        \includegraphics[width=\linewidth]{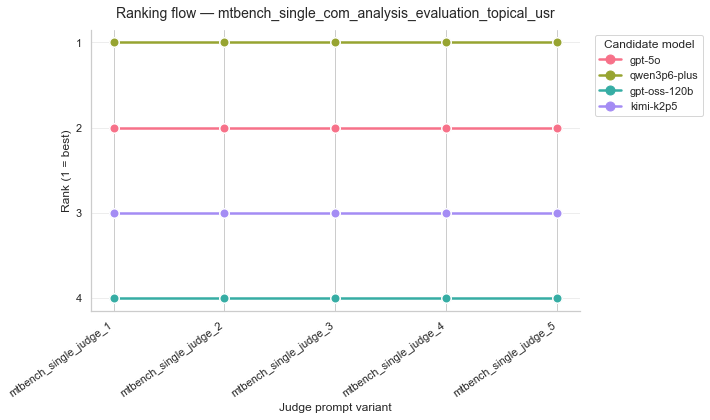}
        \caption{Topical MT Analysis.}
        \label{fig:mt_topical}
    \end{minipage}%
    \hfill
    \begin{minipage}[t]{0.48\textwidth}
        \centering
        \includegraphics[width=\linewidth]{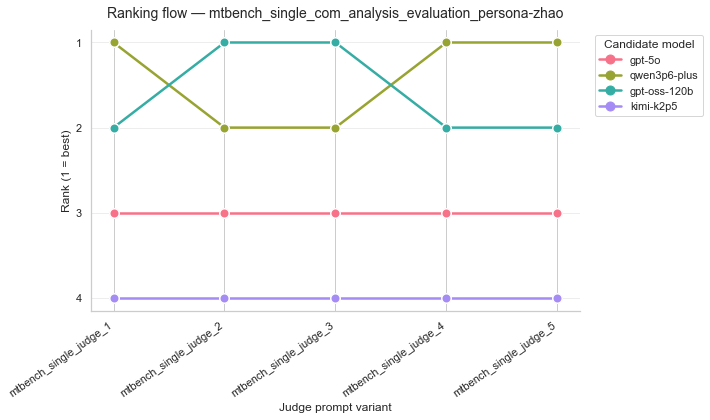}
        \caption{Persona MT Analysis.}
        \label{fig:mt_persona}
    \end{minipage}

    \vspace{0.5cm} 

    \begin{minipage}[t]{0.48\textwidth}
        \centering
        \includegraphics[width=\linewidth]{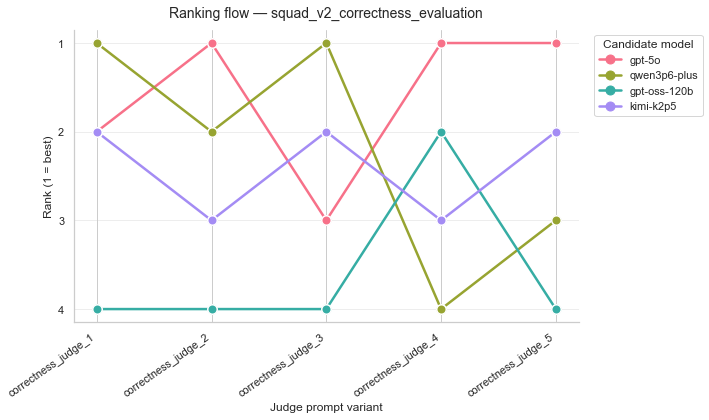}
        \caption{SQuAD 2.0 Ranking.}
        \label{fig:squad2_rank}
    \end{minipage}
\end{figure*}

Figures~~\ref{fig:mt_topical},~\ref{fig:mt_persona}, \ref{fig:squad2_rank}
and visualizes ranking trajectories across five 
paraphrastic judge prompts using bump charts. Stable evaluations produce 
parallel trajectories with no line crossings, while unstable evaluations show 
frequent crossings and changes in top-ranked systems.

The strongest Prompt Lottery effects occur in objective QA evaluation 
(SQuAD v2 correctness). Here, rankings fluctuate substantially across 
semantically equivalent prompts: GPT-4o ranks 1st under some prompts but 
drops as low as 3rd under others, while qwen3p6-plus ranges from 1st to 
4th place. Multiple reversals occur across prompts, demonstrating 
that the identity of the ``best'' model depends strongly on arbitrary prompt 
formulation.

SummEval also exhibits substantial instability, with repeated exchanges 
among the top-ranked systems and 40\% absolute reversals overall. In 
contrast, MTBench (standard dialogue evaluation) remains fully stable: 
all ranking trajectories are parallel, no crossings occur, and the same 
relative ordering is preserved across all prompt variants.

These visualizations complement the Kendall's $\tau$ statistics reported 
in Table~\ref{tab:ranking_stability} by illustrating how paraphrastic 
prompt variation changes comparative evaluation outcomes. Together, they 
show that Prompt Lottery effects are highly task-dependent, affecting 
objective evaluation substantially more than subjective dialogue assessment.

\section{Experimental setup and hyperparameters}
\label{sec:model-params}

All trainable classifiers predict binary result (whether a judge's score falls in the plurality cluster of five judges).
The model input is \texttt{LLMaJ}: the row conversation plus extracted judge rubric (\texttt{format\_model\_input}), truncated to 512 tokens (\textsc{DeBERTa} tokenizer) before sampling.
Splits are by unique instance key with no overlap across train, validation, and test.

\paragraph{Data splits (defaults).}
Up to 10{,}000 training rows (imbalanced, reflecting the holdout pool), 2{,}000 balanced validation rows, and 5{,}000 balanced test rows.

\begin{table*}[htbp]
\centering
\caption{Shared data and evaluation protocol.}
\label{tab:shared-protocol}
\small
\begin{tabular}{@{}ll@{}}
\toprule
Setting & Value \\
\midrule
Input field & \texttt{llm\_judge} (conversation + judge instruction) \\
Max.\ sequence length & 512 tokens (transformers); Llama uses char limits (below) \\
Train / val / test sizes & 10k / 2k / 5k (defaults) \\
Class weights & Inverse frequency from each model's training split \\
DeBERTa train sampling & Class-balanced from holdout pool \\
Threshold tuning (DeBERTa) & Sweep $t \in \{0.05, 0.10, \ldots, 0.95\}$ on val; metric MCC \\
Decision rule & Class~1 iff $P(y{=}1) \ge t$ (default $t{=}0.5$ unless tuned) \\
Random seed & 42 \\
\bottomrule
\end{tabular}
\end{table*}

\paragraph{Logistic regression and SVM.}
Features are mean-pooled bertbase embeddings (\texttt{bert-base-uncased}, embedding batch size 32, max length 512).
Logistic regression: \texttt{max\_iter}${=}1000$, \texttt{class\_weight}${=}$\texttt{balanced}.
SVM: RBF kernel, \texttt{probability=True}, \texttt{class\_weight}${=}$\texttt{balanced}.
Neither model is fine-tuned; both use the imbalanced 10k train split.

\paragraph{RoBERTa.}
Base model: \texttt{FacebookAI/roberta-base}\footnote{https://huggingface.co/FacebookAI/roberta-base}.
Fine-tuned for 3 epochs with learning rate $2\times10^{-5}$, batch size 16, weight decay 0.01, warmup ratio 0.1, max gradient norm 1.0, and \texttt{class\_weight} cross-entropy.
Checkpoint selection uses validation MCC each epoch on the validation split in the current implementation.
Precision: bf16 when supported, else fp16.
Input: \texttt{llm\_judge}, max length 512.

\paragraph{DeBERTa-v3-large + LoRA.}
Base model: \texttt{microsoft/deberta-v3-large}.
LoRA rank $r{=}40$, $\alpha{=}160$, dropout ${\approx}0.065$, target modules \texttt{query\_proj}, \texttt{key\_proj}, \texttt{value\_proj}.
Training: 3 epochs, learning rate ${\approx}5.25\times10^{-5}$, weight decay ${\approx}0.041$, per-device batch size 64, gradient accumulation 8 (effective batch 512), gradient checkpointing enabled, max gradient norm 0.5, warmup ratio 0.15.
Checkpoints selected by MCC on the \textbf{2k validation split}; threshold tuned on val (MCC), then metrics reported on the 5k test split.
Precision: bf16/fp16 auto-selected; no 4-bit quantization in this pipeline.

\begin{table}[htbp]
\centering
\caption{Model-specific hyperparameters (defaults).}
\label{tab:model-hparams}
\small
\begin{tabular}{@{}lcccc@{}}
\toprule
Model & Train data & LR & Batch (eff.) & Epochs \\
\midrule
Logistic regression & 10k (imbal.) & --- & --- & --- \\
SVM (RBF)           & 10k (imbal.) & --- & --- & --- \\
RoBERTa-base        & 10k (imbal.) & $2\times10^{-5}$ & 16 & 3 \\
DeBERTa + LoRA      & balanced     & $5.25\times10^{-5}$ & 64 (512) & 3 \\
Llama 3.1 8B        & none (ICL)   & --- & 64 / 4 & --- \\
\bottomrule
\end{tabular}
\end{table}


\subsection{Optuna optimization}
\begin{equation}
\tiny
  \mathrm{MCC} = \frac{TP \cdot TN - FP \cdot FN}{\sqrt{(TP + FP)(TP + FN)(TN + FP)(TN + FN)}},
\end{equation}
where $TP$, $TN$, $FP$, and $FN$ denote true positives, true negatives, false positives, and false negatives, respectively; $\mathrm{MCC} \in [-1, 1]$, with $1$ indicating perfect agreement.

We tuned DeBERTa-v3-large LoRA hyperparameters with Optuna on a single NVIDIA A100 80GB GPU, using a multi-objective study (35 trials, 2 training epochs per trial) that maximized validation Matthews correlation coefficient and balanced accuracy on an 80/20 stratified holdout split of the training export; gradient accumulation (8 steps), sequence length (512 tokens), and LoRA target modules (query, key, and value projections) were held fixed.

\begin{table}[t]
\centering
\caption{Optuna hyperparameter search space for DeBERTa LoRA.}

\begin{tabular}{llll}
\toprule
Hyperparameter & Search & Range / values & Fixed \\
\midrule
Compute & -- & -- & 1$\times$ NVIDIA A100 80GB \\
Learning rate & Log-uniform & $10^{-5}$--$2\times10^{-4}$ & -- \\
LoRA rank ($r$) & Integer & 8, 16, 24, 32, 40 & -- \\
LoRA alpha & Integer & $r$ to $4r$ (step $r$) & -- \\
LoRA dropout & Uniform & 0.0--0.15 & -- \\
Weight decay & Uniform & 0.01--0.1 & -- \\
Batch size (per device) & Categorical & 32, 50, 64, 128 & -- \\
Warmup ratio & Uniform & 0.05--0.20 & -- \\
Gradient accumulation & -- & -- & 8 \\
Epochs (per trial) & -- & -- & 2 \\
Max sequence length & -- & -- & 512 \\
Base model & -- & -- & DeBERTa-v3-large \\
LoRA targets & -- & -- & Q/K/V projections \\
\bottomrule
\end{tabular}

\end{table}

\clearpage
\subsection{Llama 3.1 8B classifier prompts}

Both zero-shot and few-shot use the same system prompt and user-turn template.
Input text is the \texttt{llm\_judge} field (conversation plus judge instruction).
Queries are truncated to 3{,}500 characters; few-shot examples to 1{,}200 characters each.
The model must output only the category digit (\texttt{0} or \texttt{1}).

\begin{tcolorbox}[promptbox, title=System prompt (shared).]
\small\ttfamily\raggedright
\begin{obeylines}
You are a classifier for LLM judge agreement analysis.

You are NOT evaluating the quality or correctness of the judge prompt text itself.

Each item describes one of five judges. The five judges use different but semantically similar prompts to score the same model answer. Each judge produces a score.

Your task: predict whether this judge's score would fall in the plurality cluster — whether at least 3 out of the 5 judges assigned the same score as this judge.

Categories:
- 0: No — this judge's score is NOT shared by at least 3 of the 5 judges
- 1: Yes — this judge's score matches at least 3 of the 5 judges (plurality agreement)

Only output the category number, nothing else.
\end{obeylines}
\end{tcolorbox}

\begin{tcolorbox}[promptbox, title=Zero-shot user turn]
\small\ttfamily\raggedright
\begin{obeylines}
Judge instruction (one of five semantically similar judge rubrics used to score the same answer):

\{llm\_judge\}

Question: Among 5 judges with different but semantically similar prompts, would this judge's score be the same as at least 3 of them?

Category:
\end{obeylines}
\end{tcolorbox}

\begin{tcolorbox}[promptbox, title=Few-shot user turn.]
\small\ttfamily\raggedright
\begin{obeylines}
{\ttfamily\small\noindent
Judge instruction (one of five semantically similar judge rubrics used to score the same answer):

\{few\_shot\_example\_1\_judge\}

Question: Among 5 judges with different but semantically similar prompts, would this judge's score be the same as at least 3 of them?

Category: 1

Judge instruction (one of five semantically similar judge rubrics used to score the same answer):

\{few\_shot\_example\_2\_judge\}

Question: Among 5 judges with different but semantically similar prompts, would this judge's score be the same as at least 3 of them?

Category: 0

Judge instruction (one of five semantically similar judge rubrics used to score the same answer):

\{llm\_judge\}

Question: Among 5 judges with different but semantically similar prompts, would this judge's score be the same as at least 3 of them?

Category:
}
\end{obeylines}
\end{tcolorbox}
\section{LLM Judge Prompt Groups} \label{appendix:prompts}

\subsection*{Organization and Navigation}
 
This appendix contains 15 groups (G1--G15) of semantically equivalent judge 
prompts, organized by evaluation task and dimension. Each group contains 4--5 
prompt variants (P1--P5) designed to be semantically equivalent while differing 
substantially in surface realization (role framing, instruction syntax, output 
format, etc.; see Table 10 for dimension taxonomy).
 
\begin{table*}[h!]
\centering
\caption{Summary of judge prompt groups. Groups G1--G5 cover 
semantic alignment, QA, and general response quality. Groups G6--G9 cover 
dialogue-specific dimensions. Groups G10--G13 cover summarization quality. 
Groups G14--G15 cover context-aware and domain-specific evaluation. $\dagger$G15 is only used
for the gated aggregation and thus not semantic variations.}
\small
\begin{tabular}{|l|l|l|}
\hline
\textbf{Group} & \textbf{Task Type} & \textbf{Prompts} \\
\hline
G1 & Semantic Alignment (Likert 1--5) & P1--P5 \\
G2 & QA Correctness (Likert 1--5) & P1--P5 \\
G3 & QA Groundedness (Likert 1--5) & P1--P5 \\
G4 & General Quality -- User Study (Likert 1--5) & P1--P5 \\
G5 & MT-Bench Quality (Likert 1--10) & P1--P5 \\
\hline
G6 & Dialogue -- Relevance (Binary) & P1--P5 \\
G7 & Dialogue -- Interesting (Binary) & P1--P5 \\
G8 & Dialogue -- Understandable (Binary) & P1--P5 \\
G9 & Dialogue -- Specific (Binary) & P1--P5 \\
\hline
G10 & SummEval -- Coherence (Likert 1--5) & P1--P5 \\
G11 & SummEval -- Consistency (Likert 1--5) & P1--P5 \\
G12 & SummEval -- Fluency (Likert 1--5) & P1--P5 \\
G13 & SummEval -- Relevance (Likert 1--5) & P1--P5 \\
\hline
G14 & Context-Aware Quality (Likert) & P1--P4 \\
\hline
G15 & Domain-Specific (Likert 1--5) & P1--P3$\dagger$ \\
\hline
\end{tabular}

\label{tab:appendix_structure}
\end{table*}
 
\subsection*{Quick Reference by Dimension}
 
\noindent\textbf{By Scoring Granularity:}
\begin{itemize}
    \item Binary (Yes/No): G6, G7, G8, G9, G14 (partial)
    \item Likert 1--5: G1, G2, G3, G4, G10, G11, G12, G13, G15
    \item Likert 1--10: G5
\end{itemize}
 
\noindent\textbf{By Task Category:}
\begin{itemize}
    \item Reference-based QA: G2, G3
    \item Semantic alignment: G1
    \item Dialogue quality: G4, G5, G6, G7, G8, G9
    \item Summarization: G10, G11, G12, G13
    \item Context-grounded QA: G14
    \item Domain-specific: G15
\end{itemize}
 
\noindent\textbf{By Paraphrastic Dimensions (see Table 10):}
\begin{itemize}
    \item Role framing variation: All groups (e.g., ``expert evaluator'' vs ``judge'')
    \item Instruction syntax: All groups (imperative vs declarative)
    \item Output format strictness: G5, G6--G9 (explicit format specifications)
    \item Task contextualization: G2, G3, G5 (verbose vs direct)
    \item Linguistic register: All groups (e.g., ``assess'' vs ``evaluate'' vs ``rate'')
\end{itemize}
 
\subsection{Prompts}

Each subsection corresponds to one group of prompts that were evaluated together for inter-judge agreement.
Within every group the five prompts share the same scoring task but differ in framing, vocabulary, or
structural layout.
Placeholder labels $\mathcal{P}_1$--$\mathcal{P}_5$ identify individual judges within a group;
placeholder group labels $\mathcal{G}_1$--$\mathcal{G}_{15}$ identify the groups.
Template variables \texttt{\{conversation\}}, \texttt{\{prediction\}}, \texttt{\{reference\}},
and \texttt{\{context\}} are filled at inference time.

\subsection*{$\mathcal{G}_1$ \quad Alignment Judges}
\textit{Task: rate (1--5) how much the prediction conveys the same information as the reference.}

\paragraph{$\mathcal{P}_1$}
\textbf{System prompt:}\newline
\begin{tcolorbox}[promptbox, title=Variant 1 ($P_1$)]
\small\ttfamily\raggedright
\begin{obeylines}
Act as a judge scoring the continuations to a conversation. You will be provided
with a conversation in the format
user: user message
assistant: assistant message.
The conversation can go for many turns. Finally you are provided with a prediction
of continuation of the conversation and a reference.
\end{obeylines}
\end{tcolorbox}
\textbf{Scoring prompt:}
\begin{tcolorbox}[promptbox, title=Variant 1 ($P_1$)]
\small\ttfamily\raggedright
\begin{obeylines}
Conversation: {conversation}

Reference: {reference}
Prediction:  {prediction}

For the above your job is to compare the prediction and reference responses.
Score whether the prediction conveys the same information as the reference on a
likert scale of 1 to 5.
Where 1 means none of the reference information is conveyed by the prediction;
and 5 means all the reference and predictions are semantically equivalent.
Output only scores from 1 to 5 (integer)
\end{obeylines}
\end{tcolorbox}

\paragraph{$\mathcal{P}_2$}
\textbf{System prompt:}\newline
\begin{tcolorbox}[promptbox, title=Variant 2 ($P_2$)]
\small\ttfamily\raggedright
\begin{obeylines}
You are an expert evaluator comparing model predictions to reference answers.
You assess how well the prediction matches the reference in terms of conveyed
information. Conversations are in the format:
user: ... assistant: ... (multiple turns allowed).
\end{obeylines}
\end{tcolorbox}
\textbf{Scoring prompt:}
\begin{tcolorbox}[promptbox, title=Variant 2 ($P_2$)]
\small\ttfamily\raggedright
\begin{obeylines}
Conversation:
{conversation}

Reference answer:
{reference}

Model prediction:
{prediction}

Task: Compare the prediction to the reference. Rate on a scale of 1 to 5 how much
of the reference's information is conveyed by the prediction.
- 1: None of the reference information is conveyed.
- 2: Little overlap.
- 3: Moderate overlap.
- 4: Most information conveyed, minor differences.
- 5: Prediction and reference are semantically equivalent.

Respond with a single integer from 1 to 5.
\end{obeylines}
\end{tcolorbox}

\paragraph{$\mathcal{P}_3$}
\textbf{System prompt:}\newline
\begin{tcolorbox}[promptbox, title=Variant 3 ($P_3$)]
\small\ttfamily\raggedright
\begin{obeylines}
You act as a rater for dialogue continuations. Given a conversation (user/assistant
turns), a reference continuation, and a model prediction, you score how closely the
prediction aligns with the reference in meaning and information.
\end{obeylines}
\end{tcolorbox}
\textbf{Scoring prompt:}
\begin{tcolorbox}[promptbox, title=Variant 3 ($P_3$)]
\small\ttfamily\raggedright
\begin{obeylines}
Dialogue context (user / assistant turns):
{conversation}

Reference continuation:
{reference}

Predicted continuation:
{prediction}

Score how well the prediction conveys the same information as the reference.
Use a 1-5 scale:
1 = prediction conveys none of the reference information
5 = prediction and reference are semantically equivalent

Output only one integer (1, 2, 3, 4, or 5).
\end{obeylines}
\end{tcolorbox}

\paragraph{$\mathcal{P}_4$}
\textbf{System prompt:}\newline
\begin{tcolorbox}[promptbox, title=Variant 4 ($P_4$)]
\small\ttfamily\raggedright
\begin{obeylines}
You are a judge evaluating whether model-generated continuations match reference
answers in content. The input includes a multi-turn conversation, a reference
response, and a predicted response. Score alignment on a 1-5 scale.
\end{obeylines}
\end{tcolorbox}
\textbf{Scoring prompt:}
\begin{tcolorbox}[promptbox, title=Variant 4 ($P_4$)]
\small\ttfamily\raggedright
\begin{obeylines}
Context (conversation):
{conversation}

Reference response:
{reference}

Predicted response:
{prediction}

Evaluate: To what degree does the prediction convey the same information as the
reference?
- 1: No overlap; prediction does not convey reference information.
- 2: Slight overlap.
- 3: About half of the reference information is conveyed.
- 4: Most reference information conveyed.
- 5: Full semantic equivalence between prediction and reference.

Reply with only the score (1-5).
\end{obeylines}
\end{tcolorbox}

\paragraph{$\mathcal{P}_5$}
\textbf{System prompt:}\newline
\begin{tcolorbox}[promptbox, title=Variant 5 ($P_5$)]
\small\ttfamily\raggedright
\begin{obeylines}
You score dialogue continuations by comparing a model's prediction to a reference
answer. The dialogue is given as user and assistant turns. Your output is a single
integer from 1 to 5 indicating how much the prediction aligns with the reference.
\end{obeylines}
\end{tcolorbox}
\textbf{Scoring prompt:}
\begin{tcolorbox}[promptbox, title=Variant 5 ($P_5$)]
\small\ttfamily\raggedright
\begin{obeylines}
Conversation (user / assistant turns):
{conversation}

Reference:
{reference}

Prediction:
{prediction}

Compare prediction and reference. Assign a score from 1 to 5:
1 = none of the reference information is in the prediction
5 = prediction and reference are semantically equivalent

Output format: one integer, 1 to 5.
\end{obeylines}
\end{tcolorbox}

\subsection*{$\mathcal{G}_2$ \quad Correctness Judges}
\textit{Task: rate (1--5) whether the prediction correctly answers the question relative to the ground-truth.}

\paragraph{$\mathcal{P}_1$}
\textbf{System prompt:}\newline
\begin{tcolorbox}[promptbox, title=Variant 1 ($P_1$)]
\small\ttfamily\raggedright
\begin{obeylines}
You are an expert evaluator assessing whether model predictions correctly answer
questions. Rate correctness on a scale of 1-5.
\end{obeylines}
\end{tcolorbox}
\textbf{Scoring prompt:}
\begin{tcolorbox}[promptbox, title=Variant 1 ($P_1$)]
\small\ttfamily\raggedright
\begin{obeylines}
You are given a question, the corresponding ground-truth answer and a prediction
from a model. Compare the "Ground-truth answer" and the "Prediction" to determine
how correctly the prediction answers the question. All information in the
ground-truth answer should be present in the prediction, including numbers and
dates. There should be no contradicting statements in the prediction. The prediction
may contain extra information. If the prediction states something as a possibility,
treat it as a definitive answer.

Question: {conversation}

Ground-truth answer: {reference}

Prediction: {prediction}

Correctness scale:
5 = Fully correct: all key information present, no contradictions
4 = Mostly correct: minor omissions or imprecision
3 = Partially correct: some key details missing or ambiguous
2 = Mostly incorrect: major omissions or contradictions
1 = Incorrect: does not answer the question or contradicts ground truth

Score (1-5):
\end{obeylines}
\end{tcolorbox}

\paragraph{$\mathcal{P}_2$}
\textbf{System prompt:}\newline
\begin{tcolorbox}[promptbox, title=Variant 2 ($P_2$)]
\small\ttfamily\raggedright
\begin{obeylines}
You are an expert answer verification system. Rate answer correctness on a scale
of 1-5.
\end{obeylines}
\end{tcolorbox}
\textbf{Scoring prompt:}
\begin{tcolorbox}[promptbox, title=Variant 2 ($P_2$)]
\small\ttfamily\raggedright
\begin{obeylines}
Given a question, its correct answer, and a model's response, rate how accurately
the model's response answers the question. The response should include all key
information from the correct answer, especially specific details like numbers and
dates. The response should not contain contradictions. Additional information beyond
the correct answer is acceptable. Treat hypothetical statements as definitive
answers.

Question: {conversation}

Correct answer: {reference}

Model's response: {prediction}

5 = Fully accurate; 4 = Mostly accurate, minor gaps; 3 = Partially accurate;
2 = Mostly inaccurate; 1 = Inaccurate or contradictory

Score (1-5):
\end{obeylines}
\end{tcolorbox}

\paragraph{$\mathcal{P}_3$}
\textbf{System prompt:}\newline
\begin{tcolorbox}[promptbox, title=Variant 3 ($P_3$)]
\small\ttfamily\raggedright
\begin{obeylines}
You are a precision answer checker. Rate correctness on a scale of 1-5.
\end{obeylines}
\end{tcolorbox}
\textbf{Scoring prompt:}
\begin{tcolorbox}[promptbox, title=Variant 3 ($P_3$)]
\small\ttfamily\raggedright
\begin{obeylines}
Your task is to rate whether a model's prediction correctly answers a given question
when compared to the ground-truth answer. Check that every piece of information in
the ground-truth answer appears in the prediction, with particular attention to
numerical values and dates. The prediction may include supplementary information.
Consider possibilities mentioned in the prediction as definite answers.

Question: {conversation}

Ground-truth answer: {reference}

Model prediction: {prediction}

5 = Complete and correct; 4 = Minor omissions; 3 = Some details missing;
2 = Major omissions or errors; 1 = Incorrect or contradictory

Score (1-5):
\end{obeylines}
\end{tcolorbox}

\paragraph{$\mathcal{P}_4$}
\textbf{System prompt:}\newline
\begin{tcolorbox}[promptbox, title=Variant 4 ($P_4$)]
\small\ttfamily\raggedright
\begin{obeylines}
You are an answer validation assistant. Rate correctness on a scale of 1-5.
\end{obeylines}
\end{tcolorbox}
\textbf{Scoring prompt:}
\begin{tcolorbox}[promptbox, title=Variant 4 ($P_4$)]
\small\ttfamily\raggedright
\begin{obeylines}
Evaluate how well a generated answer correctly addresses a question by comparing it
with the reference answer. All essential information from the reference answer should
be present in the generated answer, including precise details such as numbers and
dates. The generated answer must not contain contradictory information. Extra details
in the generated answer are permitted. Statements presented as possibilities should
be treated as definite responses.

Question: {conversation}

Reference answer: {reference}

Generated answer: {prediction}

5 = Fully correct; 4 = Mostly correct; 3 = Partially correct;
2 = Mostly incorrect; 1 = Incorrect

Score (1-5):
\end{obeylines}
\end{tcolorbox}

\paragraph{$\mathcal{P}_5$}
\textbf{System prompt:}\newline
\begin{tcolorbox}[promptbox, title=Variant 5 ($P_5$)]
\small\ttfamily\raggedright
\begin{obeylines}
You are a precise answer assessment tool. Rate correctness on a scale of 1-5.
\end{obeylines}
\end{tcolorbox}
\textbf{Scoring prompt:}
\begin{tcolorbox}[promptbox, title=Variant 5 ($P_5$)]
\small\ttfamily\raggedright
\begin{obeylines}
Determine how correct a model's answer is by comparing it against the ground-truth
answer for the given question. The model's answer should contain all information
present in the ground-truth answer, with special attention to specific details
including numbers and dates. Ensure the model's answer has no contradictions. The
model's answer may provide additional context. If the model expresses something as a
possibility, consider it a definitive response.

Question: {conversation}

Ground-truth answer: {reference}

Model's answer: {prediction}

5 = Fully correct; 4 = Mostly correct; 3 = Partially correct;
2 = Mostly incorrect; 1 = Incorrect or contradictory

Score (1-5):
\end{obeylines}
\end{tcolorbox}

\subsection*{$\mathcal{G}_3$ \quad Groundedness Judges}
\textit{Task: rate (1--5) whether the prediction is grounded in the provided evidence.}

\paragraph{$\mathcal{P}_1$}
\textbf{System prompt:}\newline
\begin{tcolorbox}[promptbox, title=Variant 1 ($P_1$)]
\small\ttfamily\raggedright
\begin{obeylines}
You are an expert evaluator assessing whether model predictions are grounded in
evidence. Rate groundedness on a scale of 1-5.
\end{obeylines}
\end{tcolorbox}
\textbf{Scoring prompt:}
\begin{tcolorbox}[promptbox, title=Variant 1 ($P_1$)]
\small\ttfamily\raggedright
\begin{obeylines}
You are given a question, the corresponding evidence and a prediction from a model.
Compare the "Prediction" and the "Evidence" to determine how well the information in
the prediction is present in the evidence or can be inferred from the evidence.

Question: {conversation}

Prediction: {prediction}

Evidence: {reference}

Groundedness scale:
5 = Fully grounded: all details supported by evidence or reasonable inference
4 = Mostly grounded: minor unsupported details
3 = Partially grounded: some details not in evidence
2 = Mostly ungrounded: major unsupported or speculative content
1 = Ungrounded: prediction not supported by evidence

Score (1-5):
\end{obeylines}
\end{tcolorbox}

\paragraph{$\mathcal{P}_2$}
\textbf{System prompt:}\newline
\begin{tcolorbox}[promptbox, title=Variant 2 ($P_2$)]
\small\ttfamily\raggedright
\begin{obeylines}
You are an expert evidence verification system. Rate groundedness on a scale of 1-5.
\end{obeylines}
\end{tcolorbox}
\textbf{Scoring prompt:}
\begin{tcolorbox}[promptbox, title=Variant 2 ($P_2$)]
\small\ttfamily\raggedright
\begin{obeylines}
Given a question, a model's prediction, and supporting evidence, rate how well the
prediction is grounded in the evidence. All information in the prediction should
either be explicitly stated in the evidence or reasonably inferable from it.

Question: {conversation}

Model's prediction: {prediction}

Supporting evidence: {reference}

5 = Fully grounded; 4 = Mostly grounded; 3 = Partially grounded;
2 = Mostly ungrounded; 1 = Ungrounded

Score (1-5):
\end{obeylines}
\end{tcolorbox}

\paragraph{$\mathcal{P}_3$}
\textbf{System prompt:}\newline
\begin{tcolorbox}[promptbox, title=Variant 3 ($P_3$)]
\small\ttfamily\raggedright
\begin{obeylines}
You are a precision groundedness evaluator. Rate on a scale of 1-5.
\end{obeylines}
\end{tcolorbox}
\textbf{Scoring prompt:}
\begin{tcolorbox}[promptbox, title=Variant 3 ($P_3$)]
\small\ttfamily\raggedright
\begin{obeylines}
Your task is to rate whether a model's prediction is grounded in the provided
evidence for a given question. Check that every piece of information in the
prediction can be found in the evidence or logically derived from it.

Question: {conversation}

Model prediction: {prediction}

Evidence passage: {reference}

5 = All details supported; 4 = Minor gaps; 3 = Some unsupported details;
2 = Major unsupported content; 1 = Not grounded in evidence

Score (1-5):
\end{obeylines}
\end{tcolorbox}

\paragraph{$\mathcal{P}_4$}
\textbf{System prompt:}\newline
\begin{tcolorbox}[promptbox, title=Variant 4 ($P_4$)]
\small\ttfamily\raggedright
\begin{obeylines}
You are an evidence-based validation assistant. Rate groundedness on a scale of 1-5.
\end{obeylines}
\end{tcolorbox}
\textbf{Scoring prompt:}
\begin{tcolorbox}[promptbox, title=Variant 4 ($P_4$)]
\small\ttfamily\raggedright
\begin{obeylines}
Evaluate how well a generated prediction is grounded in the evidence provided for a
question. The prediction's information should be either directly present in the
evidence or derivable through logical inference.

Question: {conversation}

Generated prediction: {prediction}

Reference evidence: {reference}

5 = Fully supported by evidence; 4 = Mostly supported; 3 = Partially supported;
2 = Mostly unsupported; 1 = Not supported

Score (1-5):
\end{obeylines}
\end{tcolorbox}

\paragraph{$\mathcal{P}_5$}
\textbf{System prompt:}\newline
\begin{tcolorbox}[promptbox, title=Variant 5 ($P_5$)]
\small\ttfamily\raggedright
\begin{obeylines}
You are a factual grounding assessment tool. Rate groundedness on a scale of 1-5.
\end{obeylines}
\end{tcolorbox}
\textbf{Scoring prompt:}
\begin{tcolorbox}[promptbox, title=Variant 5 ($P_5$)]
\small\ttfamily\raggedright
\begin{obeylines}
Determine how well a model's prediction is grounded in the given evidence for a
question. All information in the prediction should be traceable to the evidence
either through direct mention or reasonable inference.

Question: {conversation}

Model's prediction: {prediction}

Evidence: {reference}

5 = Fully grounded; 4 = Mostly grounded; 3 = Partially grounded;
2 = Mostly ungrounded; 1 = Ungrounded

Score (1-5):
\end{obeylines}
\end{tcolorbox}

\subsection*{$\mathcal{G}_4$ \quad Metrics User Study Judges}
\textit{Task: rate (1--5) how much the prediction conveys the same information as the reference
(shared system prompt, varied scoring framing).}

\paragraph{$\mathcal{P}_1$}
\textbf{System prompt (shared across $\mathcal{P}_1$--$\mathcal{P}_5$):}
\begin{tcolorbox}[promptbox, title=Variant 1 ($P_1$)]
\small\ttfamily\raggedright
\begin{obeylines}
Act as a judge scoring the continuations to a conversation. You will be provided
with a conversation in the format
user: user message
assistant: assistant message.
The conversation can go for many turns. Finally you are provided with a prediction
of continuation of the conversation and a reference.
\end{obeylines}
\end{tcolorbox}
\textbf{Scoring prompt:}
\begin{tcolorbox}[promptbox, title=Variant 1 ($P_1$)]
\small\ttfamily\raggedright
\begin{obeylines}
Conversation: {conversation}

Reference: {reference}

Prediction:  {prediction}

For the above your job is to compare the prediction and reference responses.
Score whether the prediction conveys the same information as the reference on a
likert scale of 1 to 5.
1 means none of the reference information is conveyed by the prediction;
and 5 means reference and prediction are semantically equivalent.
Output only scores from 1 to 5 (integer)
\end{obeylines}
\end{tcolorbox}

\paragraph{$\mathcal{P}_2$}
\textbf{Scoring prompt:}
\begin{tcolorbox}[promptbox, title=Variant 2 ($P_2$)]
\small\ttfamily\raggedright
\begin{obeylines}
Conversation:
{conversation}

Reference answer:
{reference}

Model prediction:
{prediction}

Task: Compare the prediction to the reference. Rate on a scale of 1 to 5 how much
of the reference's information is conveyed by the prediction.
- 1: None of the reference information is conveyed.
- 2: Little overlap.
- 3: Moderate overlap.
- 4: Most information conveyed, minor differences.
- 5: Prediction and reference are semantically equivalent.

Respond with a single integer from 1 to 5.
\end{obeylines}
\end{tcolorbox}

\paragraph{$\mathcal{P}_3$}
\textbf{Scoring prompt:}
\begin{tcolorbox}[promptbox, title=Variant 3 ($P_3$)]
\small\ttfamily\raggedright
\begin{obeylines}
Dialogue context (user / assistant turns):
{conversation}

Reference continuation:
{reference}

Predicted continuation:
{prediction}

Score how well the prediction conveys the same information as the reference.
Use a 1-5 scale:
1 = prediction conveys none of the reference information
5 = prediction and reference are semantically equivalent

Output only one integer (1, 2, 3, 4, or 5).
\end{obeylines}
\end{tcolorbox}

\paragraph{$\mathcal{P}_4$}
\textbf{Scoring prompt:}
\begin{tcolorbox}[promptbox, title=Variant 4 ($P_4$)]
\small\ttfamily\raggedright
\begin{obeylines}
Context (conversation):
{conversation}

Reference response:
{reference}

Predicted response:
{prediction}

Evaluate: To what degree does the prediction convey the same information as the
reference?
- 1: No overlap; prediction does not convey reference information.
- 2: Slight overlap.
- 3: About half of the reference information is conveyed.
- 4: Most reference information conveyed.
- 5: Full semantic equivalence between prediction and reference.

Reply with only the score (1-5).
\end{obeylines}
\end{tcolorbox}

\paragraph{$\mathcal{P}_5$}
\textbf{Scoring prompt:}
\begin{tcolorbox}[promptbox, title=Variant 5 ($P_5$)]
\small\ttfamily\raggedright
\begin{obeylines}
Conversation (user / assistant turns):
{conversation}

Reference:
{reference}

Prediction:
{prediction}

Compare prediction and reference. Assign a score from 1 to 5:
1 = none of the reference information is in the prediction
5 = prediction and reference are semantically equivalent

Output format: one integer, 1 to 5.
\end{obeylines}
\end{tcolorbox}

\subsection*{$\mathcal{G}_5$ \quad MT-Bench Single-Response Judges}
\textit{Task: rate (1--10) overall response quality; varied system prompt emphasis, identical scoring format.}

\paragraph{$\mathcal{P}_1$}
\textbf{System prompt:}\newline
\begin{tcolorbox}[promptbox, title=Variant 1 ($P_1$)]
\small\ttfamily\raggedright
\begin{obeylines}
Please act as an impartial judge and evaluate the quality of the response provided
by an AI assistant to the user question displayed below. Your evaluation should
consider factors such as the helpfulness, relevance, accuracy, depth, creativity,
and level of detail of the response. Begin your evaluation by providing a short
explanation. Be as objective as possible. After providing your explanation, please
rate the response on a scale of 1 to 10 by strictly following this format:
"[[rating]]", for example: "Rating: [[5]]".
\end{obeylines}
\end{tcolorbox}
\textbf{Scoring prompt (shared across $\mathcal{P}_1$--$\mathcal{P}_5$):}
\begin{tcolorbox}[promptbox, title=Variant 1 ($P_1$)]
\small\ttfamily\raggedright
\begin{obeylines}
[Question]
{conversation}
[The Start of Assistant's Answer]
{prediction}
[The End of Assistant's Answer]
\end{obeylines}
\end{tcolorbox}

\paragraph{$\mathcal{P}_2$}
\textbf{System prompt:}\newline
\begin{tcolorbox}[promptbox, title=Variant 2 ($P_2$)]
\small\ttfamily\raggedright
\begin{obeylines}
You are an impartial evaluator assessing AI assistant responses. Consider coherence,
informativeness, and how well the response addresses the user's needs. Provide a
brief explanation, then rate 1-10 using exactly: Rating: [[N]]
\end{obeylines}
\end{tcolorbox}

\paragraph{$\mathcal{P}_3$}
\textbf{System prompt:}\newline
\begin{tcolorbox}[promptbox, title=Variant 3 ($P_3$)]
\small\ttfamily\raggedright
\begin{obeylines}
Act as an objective judge evaluating AI responses. Assess accuracy, completeness,
and whether the answer fully addresses the question. Give a short justification,
then rate 1-10 in the format: Rating: [[N]]
\end{obeylines}
\end{tcolorbox}

\paragraph{$\mathcal{P}_4$}
\textbf{System prompt:}\newline
\begin{tcolorbox}[promptbox, title=Variant 4 ($P_4$)]
\small\ttfamily\raggedright
\begin{obeylines}
You are a neutral judge rating AI assistant responses. Evaluate clarity,
usefulness, and appropriateness for the context. Explain briefly, then provide
your rating 1-10 as: Rating: [[N]]
\end{obeylines}
\end{tcolorbox}

\paragraph{$\mathcal{P}_5$}
\textbf{System prompt:}\newline
\begin{tcolorbox}[promptbox, title=Variant 5 ($P_5$)]
\small\ttfamily\raggedright
\begin{obeylines}
Act as an impartial judge. Holistically evaluate the AI response for overall
quality, considering relevance, helpfulness, and user satisfaction. Provide a
concise explanation, then rate 1-10 using: Rating: [[N]]
\end{obeylines}
\end{tcolorbox}

\subsection*{$\mathcal{G}_6$ \quad Relevance Judges (Dialogue)}
\textit{Task: binary (Yes/No) whether the response is relevant to the conversation context.}

\paragraph{$\mathcal{P}_1$}
\textbf{System prompt:}\newline
\begin{tcolorbox}[promptbox, title=Variant 1 ($P_1$)]
\small\ttfamily\raggedright
\begin{obeylines}
You are an expert evaluator who assesses whether responses are relevant to their
conversation context.
\end{obeylines}
\end{tcolorbox}
\textbf{Scoring prompt:}
\begin{tcolorbox}[promptbox, title=Variant 1 ($P_1$)]
\small\ttfamily\raggedright
\begin{obeylines}
\#\#\# Context:

{conversation}

\#\#\# Response:

{prediction}

\#\#\# Instruction:

Above is a dialogue context and the corresponding response.

Question: Is the response relevant to the context?

\#\#\# Your Answer (respond with ONLY 'Yes' or 'No'):
\end{obeylines}
\end{tcolorbox}

\paragraph{$\mathcal{P}_2$}
\textbf{System prompt:}\newline
\begin{tcolorbox}[promptbox, title=Variant 2 ($P_2$)]
\small\ttfamily\raggedright
\begin{obeylines}
You are an expert evaluator specializing in assessing response appropriateness in
conversations.
\end{obeylines}
\end{tcolorbox}
\textbf{Scoring prompt:}
\begin{tcolorbox}[promptbox, title=Variant 2 ($P_2$)]
\small\ttfamily\raggedright
\begin{obeylines}
Conversation Context:
{conversation}

Given Response:
{prediction}

Task: Determine if the given response is appropriate and relevant to the
conversation context.

Does this response appropriately address or relate to the preceding conversation?

Respond with ONLY 'Yes' (if relevant) or 'No' (if not relevant):
\end{obeylines}
\end{tcolorbox}

\paragraph{$\mathcal{P}_3$}
\textbf{System prompt:}\newline
\begin{tcolorbox}[promptbox, title=Variant 3 ($P_3$)]
\small\ttfamily\raggedright
\begin{obeylines}
You are a precise evaluator who determines if conversational responses maintain
proper connection to their context.
\end{obeylines}
\end{tcolorbox}
\textbf{Scoring prompt:}
\begin{tcolorbox}[promptbox, title=Variant 3 ($P_3$)]
\small\ttfamily\raggedright
\begin{obeylines}
Review the following:

Dialogue Context: {conversation}

Response to Evaluate: {prediction}

Evaluation Question: Does the response maintain a logical connection to the
dialogue context?

IMPORTANT: Answer with ONLY 'Yes' or 'No':
- 'Yes' if the response is connected and relevant
- 'No' if the response is disconnected or irrelevant

Your answer:
\end{obeylines}
\end{tcolorbox}

\paragraph{$\mathcal{P}_4$}
\textbf{System prompt:}\newline
\begin{tcolorbox}[promptbox, title=Variant 4 ($P_4$)]
\small\ttfamily\raggedright
\begin{obeylines}
You are an expert judge evaluating whether conversational responses stay on-topic.
\end{obeylines}
\end{tcolorbox}
\textbf{Scoring prompt:}
\begin{tcolorbox}[promptbox, title=Variant 4 ($P_4$)]
\small\ttfamily\raggedright
\begin{obeylines}
Context of the conversation:
{conversation}

Actual response:
{prediction}

Assessment: Is the actual response on-topic and relevant to the conversation
context?

Provide your evaluation:
- 'Yes' if the response is relevant and on-topic
- 'No' if the response is irrelevant or off-topic

Answer (Yes or No only):
\end{obeylines}
\end{tcolorbox}

\paragraph{$\mathcal{P}_5$}
\textbf{System prompt:}\newline
\begin{tcolorbox}[promptbox, title=Variant 5 ($P_5$)]
\small\ttfamily\raggedright
\begin{obeylines}
You are a binary classifier that determines response relevance in conversations.
\end{obeylines}
\end{tcolorbox}
\textbf{Scoring prompt:}
\begin{tcolorbox}[promptbox, title=Variant 5 ($P_5$)]
\small\ttfamily\raggedright
\begin{obeylines}
INPUT CONVERSATION:
{conversation}

RESPONSE TO EVALUATE:
{prediction}

CLASSIFICATION TASK:
Classify whether the response is relevant to the input conversation.

OUTPUT FORMAT: Answer with ONLY 'Yes' or 'No'
- Output 'Yes' if relevant
- Output 'No' if not relevant

CLASSIFICATION:
\end{obeylines}
\end{tcolorbox}

\subsection*{$\mathcal{G}_7$ \quad Interesting Judges}
\textit{Task: binary (Yes/No) whether the response is interesting in the conversation context.}

\paragraph{$\mathcal{P}_1$}
\textbf{System prompt:}\newline
\begin{tcolorbox}[promptbox, title=Variant 1 ($P_1$)]
\small\ttfamily\raggedright
\begin{obeylines}
You are an expert evaluator who assesses whether responses are interesting to their
conversation context.
\end{obeylines}
\end{tcolorbox}
\textbf{Scoring prompt:}
\begin{tcolorbox}[promptbox, title=Variant 1 ($P_1$)]
\small\ttfamily\raggedright
\begin{obeylines}
\#\#\# Context:

{conversation}

\#\#\# Response:

{prediction}

\#\#\# Instruction:

Above is a dialogue context and the corresponding response.

Question: Is the response interesting?

\#\#\# Your Answer (respond with ONLY 'Yes' or 'No'):
\end{obeylines}
\end{tcolorbox}

\paragraph{$\mathcal{P}_2$}
\textbf{System prompt:}\newline
\begin{tcolorbox}[promptbox, title=Variant 2 ($P_2$)]
\small\ttfamily\raggedright
\begin{obeylines}
You are an expert evaluator specializing in assessing response engagement and
interest in conversations.
\end{obeylines}
\end{tcolorbox}
\textbf{Scoring prompt:}
\begin{tcolorbox}[promptbox, title=Variant 2 ($P_2$)]
\small\ttfamily\raggedright
\begin{obeylines}
Conversation Context:
{conversation}

Given Response:
{prediction}

Task: Determine if the given response is engaging and interesting in the
conversation context.

Does this response provide interesting and engaging content?

Respond with ONLY 'Yes' (if interesting) or 'No' (if not interesting):
\end{obeylines}
\end{tcolorbox}

\paragraph{$\mathcal{P}_3$}
\textbf{System prompt:}\newline
\begin{tcolorbox}[promptbox, title=Variant 3 ($P_3$)]
\small\ttfamily\raggedright
\begin{obeylines}
You are a precise evaluator who determines if conversational responses are
captivating and interesting.
\end{obeylines}
\end{tcolorbox}
\textbf{Scoring prompt:}
\begin{tcolorbox}[promptbox, title=Variant 3 ($P_3$)]
\small\ttfamily\raggedright
\begin{obeylines}
Review the following:

Dialogue Context: {conversation}

Response to Evaluate: {prediction}

Evaluation Question: Is the response captivating and interesting?

IMPORTANT: Answer with ONLY 'Yes' or 'No':
- 'Yes' if the response is interesting and captivating
- 'No' if the response is boring or uninteresting

Your answer:
\end{obeylines}
\end{tcolorbox}

\paragraph{$\mathcal{P}_4$}
\textbf{System prompt:}\newline
\begin{tcolorbox}[promptbox, title=Variant 4 ($P_4$)]
\small\ttfamily\raggedright
\begin{obeylines}
You are an expert judge evaluating whether conversational responses are compelling
and interesting.
\end{obeylines}
\end{tcolorbox}
\textbf{Scoring prompt:}
\begin{tcolorbox}[promptbox, title=Variant 4 ($P_4$)]
\small\ttfamily\raggedright
\begin{obeylines}
Context of the conversation:
{conversation}

Actual response:
{prediction}

Assessment: Is the actual response interesting and compelling in the conversation
context?

Provide your evaluation:
- 'Yes' if the response is interesting and compelling
- 'No' if the response is dull or uninteresting

Answer (Yes or No only):
\end{obeylines}
\end{tcolorbox}

\paragraph{$\mathcal{P}_5$}
\textbf{System prompt:}\newline
\begin{tcolorbox}[promptbox, title=Variant 5 ($P_5$)]
\small\ttfamily\raggedright
\begin{obeylines}
You are a binary classifier that determines if responses are interesting in
conversations.
\end{obeylines}
\end{tcolorbox}
\textbf{Scoring prompt:}
\begin{tcolorbox}[promptbox, title=Variant 5 ($P_5$)]
\small\ttfamily\raggedright
\begin{obeylines}
INPUT CONVERSATION:
{conversation}

RESPONSE TO EVALUATE:
{prediction}

CLASSIFICATION TASK:
Classify whether the response is interesting in the context of the conversation.

OUTPUT FORMAT: Answer with ONLY 'Yes' or 'No'
- Output 'Yes' if interesting
- Output 'No' if not interesting

CLASSIFICATION:
\end{obeylines}
\end{tcolorbox}

\subsection*{$\mathcal{G}_8$ \quad Understandable Judges}
\textit{Task: binary (Yes/No) whether the response is understandable in the conversation context.}

\paragraph{$\mathcal{P}_1$}
\textbf{System prompt:}\newline
\begin{tcolorbox}[promptbox, title=Variant 1 ($P_1$)]
\small\ttfamily\raggedright
\begin{obeylines}
You are an expert evaluator who assesses whether responses are understandable in
their conversation context.
\end{obeylines}
\end{tcolorbox}
\textbf{Scoring prompt:}
\begin{tcolorbox}[promptbox, title=Variant 1 ($P_1$)]
\small\ttfamily\raggedright
\begin{obeylines}
\#\#\# Context:

{conversation}

\#\#\# Response:

{prediction}

\#\#\# Instruction:

Above is a dialogue context and the corresponding response.

Question: Is the response understandable?

\#\#\# Your Answer (respond with ONLY 'Yes' or 'No'):
\end{obeylines}
\end{tcolorbox}

\paragraph{$\mathcal{P}_2$}
\textbf{System prompt:}\newline
\begin{tcolorbox}[promptbox, title=Variant 2 ($P_2$)]
\small\ttfamily\raggedright
\begin{obeylines}
You are an expert evaluator specializing in assessing response clarity and
comprehensibility in conversations.
\end{obeylines}
\end{tcolorbox}
\textbf{Scoring prompt:}
\begin{tcolorbox}[promptbox, title=Variant 2 ($P_2$)]
\small\ttfamily\raggedright
\begin{obeylines}
Conversation Context:
{conversation}

Given Response:
{prediction}

Task: Determine if the given response is clear and understandable in the
conversation context.

Is this response clear and easy to comprehend?

Respond with ONLY 'Yes' (if understandable) or 'No' (if not understandable):
\end{obeylines}
\end{tcolorbox}

\paragraph{$\mathcal{P}_3$}
\textbf{System prompt:}\newline
\begin{tcolorbox}[promptbox, title=Variant 3 ($P_3$)]
\small\ttfamily\raggedright
\begin{obeylines}
You are a precise evaluator who determines if conversational responses are
comprehensible and understandable.
\end{obeylines}
\end{tcolorbox}
\textbf{Scoring prompt:}
\begin{tcolorbox}[promptbox, title=Variant 3 ($P_3$)]
\small\ttfamily\raggedright
\begin{obeylines}
Review the following:

Dialogue Context: {conversation}

Response to Evaluate: {prediction}

Evaluation Question: Is the response comprehensible and understandable?

IMPORTANT: Answer with ONLY 'Yes' or 'No':
- 'Yes' if the response is clear and understandable
- 'No' if the response is confusing or incomprehensible

Your answer:
\end{obeylines}
\end{tcolorbox}

\paragraph{$\mathcal{P}_4$}
\textbf{System prompt:}\newline
\begin{tcolorbox}[promptbox, title=Variant 4 ($P_4$)]
\small\ttfamily\raggedright
\begin{obeylines}
You are an expert judge evaluating whether conversational responses are
comprehensible and understandable.
\end{obeylines}
\end{tcolorbox}
\textbf{Scoring prompt:}
\begin{tcolorbox}[promptbox, title=Variant 4 ($P_4$)]
\small\ttfamily\raggedright
\begin{obeylines}
Context of the conversation:
{conversation}

Actual response:
{prediction}

Assessment: Is the actual response comprehensible and understandable in the
conversation context?

Provide your evaluation:
- 'Yes' if the response is clear and understandable
- 'No' if the response is unclear or confusing

Answer (Yes or No only):
\end{obeylines}
\end{tcolorbox}

\paragraph{$\mathcal{P}_5$}
\textbf{System prompt:}\newline
\begin{tcolorbox}[promptbox, title=Variant 5 ($P_5$)]
\small\ttfamily\raggedright
\begin{obeylines}
You are a binary classifier that determines if responses are understandable in
conversations.
\end{obeylines}
\end{tcolorbox}
\textbf{Scoring prompt:}
\begin{tcolorbox}[promptbox, title=Variant 5 ($P_5$)]
\small\ttfamily\raggedright
\begin{obeylines}
INPUT CONVERSATION:
{conversation}

RESPONSE TO EVALUATE:
{prediction}

CLASSIFICATION TASK:
Classify whether the response is understandable in the context of the conversation.

OUTPUT FORMAT: Answer with ONLY 'Yes' or 'No'
- Output 'Yes' if understandable
- Output 'No' if not understandable

CLASSIFICATION:
\end{obeylines}
\end{tcolorbox}

\subsection*{$\mathcal{G}_9$ \quad Specific Judges}
\textit{Task: binary (Yes/No) whether the response is specific (rather than generic) to the conversation context.}

\paragraph{$\mathcal{P}_1$}
\textbf{System prompt:}\newline
\begin{tcolorbox}[promptbox, title=Variant 1 ($P_1$)]
\small\ttfamily\raggedright
\begin{obeylines}
You are an expert evaluator who assesses whether responses are specific to their
conversation context.
\end{obeylines}
\end{tcolorbox}
\textbf{Scoring prompt:}
\begin{tcolorbox}[promptbox, title=Variant 1 ($P_1$)]
\small\ttfamily\raggedright
\begin{obeylines}
\#\#\# Context:

{conversation}

\#\#\# Response:

{prediction}

\#\#\# Instruction:

Above is a dialogue context and the corresponding response.

Question: Is the response specific to the context?

\#\#\# Your Answer (respond with ONLY 'Yes' or 'No'):
\end{obeylines}
\end{tcolorbox}

\paragraph{$\mathcal{P}_2$}
\textbf{System prompt:}\newline
\begin{tcolorbox}[promptbox, title=Variant 2 ($P_2$)]
\small\ttfamily\raggedright
\begin{obeylines}
You are an expert evaluator specializing in assessing whether responses are
tailored and specific to conversations.
\end{obeylines}
\end{tcolorbox}
\textbf{Scoring prompt:}
\begin{tcolorbox}[promptbox, title=Variant 2 ($P_2$)]
\small\ttfamily\raggedright
\begin{obeylines}
Conversation Context:
{conversation}

Given Response:
{prediction}

Task: Determine if the given response is tailored and specific to the conversation
context.

Is this response specific and tailored to the particular conversation context?

Respond with ONLY 'Yes' (if specific) or 'No' (if not specific):
\end{obeylines}
\end{tcolorbox}

\paragraph{$\mathcal{P}_3$}
\textbf{System prompt:}\newline
\begin{tcolorbox}[promptbox, title=Variant 3 ($P_3$)]
\small\ttfamily\raggedright
\begin{obeylines}
You are a precise evaluator who determines if conversational responses are
context-specific rather than generic.
\end{obeylines}
\end{tcolorbox}
\textbf{Scoring prompt:}
\begin{tcolorbox}[promptbox, title=Variant 3 ($P_3$)]
\small\ttfamily\raggedright
\begin{obeylines}
Review the following:

Dialogue Context: {conversation}

Response to Evaluate: {prediction}

Evaluation Question: Is the response specific to this particular context rather
than being generic?

IMPORTANT: Answer with ONLY 'Yes' or 'No':
- 'Yes' if the response is specific and context-tailored
- 'No' if the response is generic or non-specific

Your answer:
\end{obeylines}
\end{tcolorbox}

\paragraph{$\mathcal{P}_4$}
\textbf{System prompt:}\newline
\begin{tcolorbox}[promptbox, title=Variant 4 ($P_4$)]
\small\ttfamily\raggedright
\begin{obeylines}
You are an expert judge evaluating whether conversational responses are specifically
tailored to the context.
\end{obeylines}
\end{tcolorbox}
\textbf{Scoring prompt:}
\begin{tcolorbox}[promptbox, title=Variant 4 ($P_4$)]
\small\ttfamily\raggedright
\begin{obeylines}
Context of the conversation:
{conversation}

Actual response:
{prediction}

Assessment: Is the actual response specific to the conversation context rather
than being generic?

Provide your evaluation:
- 'Yes' if the response is specific and context-tailored
- 'No' if the response is generic or not specific

Answer (Yes or No only):
\end{obeylines}
\end{tcolorbox}

\paragraph{$\mathcal{P}_5$}
\textbf{System prompt:}\newline
\begin{tcolorbox}[promptbox, title=Variant 5 ($P_5$)]
\small\ttfamily\raggedright
\begin{obeylines}
You are a binary classifier that determines if responses are specific to the
conversation context.
\end{obeylines}
\end{tcolorbox}
\textbf{Scoring prompt:}
\begin{tcolorbox}[promptbox, title=Variant 5 ($P_5$)]
\small\ttfamily\raggedright
\begin{obeylines}
INPUT CONVERSATION:
{conversation}

RESPONSE TO EVALUATE:
{prediction}

CLASSIFICATION TASK:
Classify whether the response is specific to the conversation context or generic.

OUTPUT FORMAT: Answer with ONLY 'Yes' or 'No'
- Output 'Yes' if specific to context
- Output 'No' if generic or not specific

CLASSIFICATION:
\end{obeylines}
\end{tcolorbox}

\subsection*{$\mathcal{G}_{10}$ \quad SummEval Coherence Judges}
\textit{Task: rate (1--5) the structural coherence and logical flow of a summary.}

\paragraph{$\mathcal{P}_1$}
\textbf{System prompt:}\newline
\begin{tcolorbox}[promptbox, title=Variant 1 ($P_1$)]
\small\ttfamily\raggedright
\begin{obeylines}
You are an expert evaluator assessing the coherence of text summaries. Rate
summaries on a scale of 1-5 based on their structure and organization.
\end{obeylines}
\end{tcolorbox}
\textbf{Scoring prompt:}
\begin{tcolorbox}[promptbox, title=Variant 1 ($P_1$)]
\small\ttfamily\raggedright
\begin{obeylines}
You will be given one summary written for a source document.

Your task is to rate the summary on one metric: Coherence.

Please make sure you read and understand these instructions carefully. Please keep
this document open while reviewing, and refer to it as needed.

Evaluation Criteria:

Coherence (1-5) - the collective quality of all sentences. We align this dimension
with the DUC quality question of structure and coherence whereby "the summary should
be well-structured and well-organized. The summary should not just be a heap of
related information, but should build from sentence to sentence to a coherent body
of information about a topic."

Evaluation Steps:

1. Read the source document carefully.
2. Read the summary and compare it to the source document.
3. Assign a score for coherence on a scale of 1-5, where 1 is the lowest and 5 is
   the highest.

Source Document:
{conversation}

Summary:
{prediction}

Evaluation Form (scores ONLY):

- Coherence (1-5):
\end{obeylines}
\end{tcolorbox}

\paragraph{$\mathcal{P}_2$}
\textbf{System prompt:}\newline
\begin{tcolorbox}[promptbox, title=Variant 2 ($P_2$)]
\small\ttfamily\raggedright
\begin{obeylines}
You are an expert at evaluating summary organization. Assess how well-structured
and logically organized the summary is.
\end{obeylines}
\end{tcolorbox}
\textbf{Scoring prompt:}
\begin{tcolorbox}[promptbox, title=Variant 2 ($P_2$)]
\small\ttfamily\raggedright
\begin{obeylines}
Evaluate the following summary's organizational quality and structural coherence.

Source Text:
{conversation}

Summary to Evaluate:
{prediction}

Assessment Criteria:
Rate the summary's organization and structure on a scale of 1 to 5:

5 - Excellently organized with clear logical flow
4 - Well-structured with good organization
3 - Adequately organized with some logical flow
2 - Poorly organized with limited structure
1 - Incoherent with no clear organization

Consider:
- Does the summary build logically from sentence to sentence?
- Is information presented in a coherent, structured manner?
- Does it avoid being just a heap of disconnected facts?

Your Rating (1-5):
\end{obeylines}
\end{tcolorbox}

\paragraph{$\mathcal{P}_3$}
\textbf{System prompt:}\newline
\begin{tcolorbox}[promptbox, title=Variant 3 ($P_3$)]
\small\ttfamily\raggedright
\begin{obeylines}
You are a summary quality expert focusing on logical flow and coherent structure.
\end{obeylines}
\end{tcolorbox}
\textbf{Scoring prompt:}
\begin{tcolorbox}[promptbox, title=Variant 3 ($P_3$)]
\small\ttfamily\raggedright
\begin{obeylines}
Assess the coherence of this summary by evaluating its logical flow and structural
quality.

Original Document:
{conversation}

Summary:
{prediction}

Coherence Evaluation Guidelines:
- Score 5: Exceptional logical flow, perfectly structured
- Score 4: Strong coherence, well-organized progression
- Score 3: Moderate coherence, acceptable structure
- Score 2: Weak coherence, disjointed organization
- Score 1: No coherence, random collection of sentences

Question: How coherent is this summary in terms of structure and logical
progression?

Score (1-5):
\end{obeylines}
\end{tcolorbox}

\paragraph{$\mathcal{P}_4$}
\textbf{System prompt:}\newline
\begin{tcolorbox}[promptbox, title=Variant 4 ($P_4$)]
\small\ttfamily\raggedright
\begin{obeylines}
You are an evaluator comparing summary structure and coherence against reference
summaries.
\end{obeylines}
\end{tcolorbox}
\textbf{Scoring prompt:}
\begin{tcolorbox}[promptbox, title=Variant 4 ($P_4$)]
\small\ttfamily\raggedright
\begin{obeylines}
Compare the structural coherence of the generated summary.

Source:
{conversation}

Generated Summary:
{prediction}

Rate the generated summary's coherence (structure and organization):

1 = Completely incoherent
2 = Mostly disorganized
3 = Somewhat coherent
4 = Mostly coherent
5 = Highly coherent and well-organized

Your coherence score (1-5):
\end{obeylines}
\end{tcolorbox}

\paragraph{$\mathcal{P}_5$}
\textbf{System prompt:}\newline
\begin{tcolorbox}[promptbox, title=Variant 5 ($P_5$)]
\small\ttfamily\raggedright
\begin{obeylines}
You are a coherence classifier for text summaries, rating organization quality on
a 5-point scale.
\end{obeylines}
\end{tcolorbox}
\textbf{Scoring prompt:}
\begin{tcolorbox}[promptbox, title=Variant 5 ($P_5$)]
\small\ttfamily\raggedright
\begin{obeylines}
DOCUMENT:
{conversation}

SUMMARY:
{prediction}

TASK: Rate the summary's coherence (organizational quality and logical structure)

SCALE:
1 - Incoherent/Disorganized
2 - Poorly Structured
3 - Adequately Organized
4 - Well-Structured
5 - Excellently Coherent

COHERENCE RATING:
\end{obeylines}
\end{tcolorbox}

\subsection*{$\mathcal{G}_{11}$ \quad SummEval Consistency Judges}
\textit{Task: rate (1--5) factual consistency of a summary with the source document.}

\paragraph{$\mathcal{P}_1$}
\textbf{System prompt:}\newline
\begin{tcolorbox}[promptbox, title=Variant 1 ($P_1$)]
\small\ttfamily\raggedright
\begin{obeylines}
You are an expert evaluator assessing the factual consistency of text summaries.
Rate summaries on a scale of 1-5 based on their factual alignment with the source.
\end{obeylines}
\end{tcolorbox}
\textbf{Scoring prompt:}
\begin{tcolorbox}[promptbox, title=Variant 1 ($P_1$)]
\small\ttfamily\raggedright
\begin{obeylines}
You will be given one summary written for a source document.

Your task is to rate the summary on one metric: Consistency.

Please make sure you read and understand these instructions carefully. Please keep
this document open while reviewing, and refer to it as needed.

Evaluation Criteria:

Consistency (1-5) - the factual alignment between the summary and the summarized
source. A factually consistent summary contains only statements that are entailed
by the source document. Annotators were also asked to penalize summaries that
contained hallucinated facts.

Evaluation Steps:

1. Read the source document carefully.
2. Read the summary and compare it to the source document.
3. Assign a score for consistency on a scale of 1-5, where 1 is the lowest and 5
   is the highest.

Source Document:
{conversation}

Summary:
{prediction}

Evaluation Form (scores ONLY):

- Consistency (1-5):
\end{obeylines}
\end{tcolorbox}

\paragraph{$\mathcal{P}_2$}
\textbf{System prompt:}\newline
\begin{tcolorbox}[promptbox, title=Variant 2 ($P_2$)]
\small\ttfamily\raggedright
\begin{obeylines}
You are an expert at evaluating factual consistency. Assess how accurately the
summary reflects the source document.
\end{obeylines}
\end{tcolorbox}
\textbf{Scoring prompt:}
\begin{tcolorbox}[promptbox, title=Variant 2 ($P_2$)]
\small\ttfamily\raggedright
\begin{obeylines}
Evaluate the following summary's factual consistency with its source.

Source Text:
{conversation}

Summary to Evaluate:
{prediction}

Assessment Criteria:
Rate the summary's factual consistency on a scale of 1 to 5:

5 - Perfectly consistent, all facts align with source
4 - Mostly consistent, minor inaccuracies
3 - Moderately consistent, some factual issues
2 - Inconsistent, multiple factual errors
1 - Highly inconsistent, contains hallucinations

Consider:
- Does the summary contain only facts from the source?
- Are there any hallucinated or invented details?
- Does it misrepresent any information?

Your Rating (1-5):
\end{obeylines}
\end{tcolorbox}

\paragraph{$\mathcal{P}_3$}
\textbf{System prompt:}\newline
\begin{tcolorbox}[promptbox, title=Variant 3 ($P_3$)]
\small\ttfamily\raggedright
\begin{obeylines}
You are a summary quality expert focusing on factual entailment and consistency
verification.
\end{obeylines}
\end{tcolorbox}
\textbf{Scoring prompt:}
\begin{tcolorbox}[promptbox, title=Variant 3 ($P_3$)]
\small\ttfamily\raggedright
\begin{obeylines}
Assess the factual consistency of this summary by verifying all statements are
entailed by the source.

Original Document:
{conversation}

Summary:
{prediction}

Consistency Evaluation Guidelines:
- Score 5: All statements fully entailed by source
- Score 4: Nearly all statements supported, minimal issues
- Score 3: Most statements consistent, some unsupported claims
- Score 2: Many statements not supported, factual errors present
- Score 1: Contains hallucinations, contradicts source

Question: How factually consistent is this summary with the source document?

Score (1-5):
\end{obeylines}
\end{tcolorbox}

\paragraph{$\mathcal{P}_4$}
\textbf{System prompt:}\newline
\begin{tcolorbox}[promptbox, title=Variant 4 ($P_4$)]
\small\ttfamily\raggedright
\begin{obeylines}
You are an evaluator comparing factual consistency of summaries against source
documents and references.
\end{obeylines}
\end{tcolorbox}
\textbf{Scoring prompt:}
\begin{tcolorbox}[promptbox, title=Variant 4 ($P_4$)]
\small\ttfamily\raggedright
\begin{obeylines}
Compare the factual consistency of the generated summary with the source.

Source:
{conversation}

Reference Summary:
{reference}

Generated Summary:
{prediction}

Rate the generated summary's consistency (factual alignment with source):

1 = Contains major hallucinations/contradictions
2 = Multiple factual inaccuracies
3 = Somewhat factually aligned
4 = Mostly factually accurate
5 = Perfectly consistent with source

Your consistency score (1-5):
\end{obeylines}
\end{tcolorbox}

\paragraph{$\mathcal{P}_5$}
\textbf{System prompt:}\newline
\begin{tcolorbox}[promptbox, title=Variant 5 ($P_5$)]
\small\ttfamily\raggedright
\begin{obeylines}
You are a factual consistency classifier for text summaries, rating accuracy on a
5-point scale.
\end{obeylines}
\end{tcolorbox}
\textbf{Scoring prompt:}
\begin{tcolorbox}[promptbox, title=Variant 5 ($P_5$)]
\small\ttfamily\raggedright
\begin{obeylines}
DOCUMENT:
{conversation}

SUMMARY:
{prediction}

TASK: Rate the summary's consistency (factual alignment with source)

SCALE:
1 - Highly Inconsistent/Hallucinations
2 - Multiple Factual Errors
3 - Moderately Consistent
4 - Mostly Accurate
5 - Perfectly Consistent

CONSISTENCY RATING:
\end{obeylines}
\end{tcolorbox}

\subsection*{$\mathcal{G}_{12}$ \quad SummEval Fluency Judges}
\textit{Task: rate (1--5) the grammatical quality and readability of a summary.}

\paragraph{$\mathcal{P}_1$}
\textbf{System prompt:}\newline
\begin{tcolorbox}[promptbox, title=Variant 1 ($P_1$)]
\small\ttfamily\raggedright
\begin{obeylines}
You are an expert evaluator assessing the fluency of text summaries. Rate summaries
on a scale of 1-5 based on their grammar and readability.
\end{obeylines}
\end{tcolorbox}
\textbf{Scoring prompt:}
\begin{tcolorbox}[promptbox, title=Variant 1 ($P_1$)]
\small\ttfamily\raggedright
\begin{obeylines}
You will be given one summary written for a source document.

Your task is to rate the summary on one metric: Fluency.

Please make sure you read and understand these instructions carefully. Please keep
this document open while reviewing, and refer to it as needed.

Evaluation Criteria:

Fluency (1-5) - the quality of individual sentences. The sentences should have no
formatting problems, capitalization errors or obviously ungrammatical sentences
(e.g., fragments, missing components) that make the text difficult to read.

Evaluation Steps:

1. Read the summary carefully.
2. Evaluate the quality of the writing.
3. Assign a score for fluency on a scale of 1-5, where 1 is the lowest and 5 is
   the highest.

Source Document:
{conversation}

Summary:
{prediction}

Evaluation Form (scores ONLY):

- Fluency (1-5):
\end{obeylines}
\end{tcolorbox}

\paragraph{$\mathcal{P}_2$}
\textbf{System prompt:}\newline
\begin{tcolorbox}[promptbox, title=Variant 2 ($P_2$)]
\small\ttfamily\raggedright
\begin{obeylines}
You are an expert at evaluating text quality. Assess the grammatical correctness
and readability of the summary.
\end{obeylines}
\end{tcolorbox}
\textbf{Scoring prompt:}
\begin{tcolorbox}[promptbox, title=Variant 2 ($P_2$)]
\small\ttfamily\raggedright
\begin{obeylines}
Evaluate the following summary's fluency and grammatical quality.

Source Text:
{conversation}

Summary to Evaluate:
{prediction}

Assessment Criteria:
Rate the summary's fluency on a scale of 1 to 5:

5 - Perfectly fluent, no grammatical errors, excellent readability
4 - Mostly fluent, minor grammatical issues
3 - Adequately fluent, some grammar or readability issues
2 - Poor fluency, multiple grammatical errors
1 - Very poor fluency, difficult to read

Consider:
- Are all sentences grammatically correct?
- Is the text easy to read and understand?
- Are there formatting or capitalization problems?

Your Rating (1-5):
\end{obeylines}
\end{tcolorbox}

\paragraph{$\mathcal{P}_3$}
\textbf{System prompt:}\newline
\begin{tcolorbox}[promptbox, title=Variant 3 ($P_3$)]
\small\ttfamily\raggedright
\begin{obeylines}
You are a summary quality expert focusing on writing quality and grammatical
correctness.
\end{obeylines}
\end{tcolorbox}
\textbf{Scoring prompt:}
\begin{tcolorbox}[promptbox, title=Variant 3 ($P_3$)]
\small\ttfamily\raggedright
\begin{obeylines}
Assess the fluency of this summary by evaluating its grammatical quality and
readability.

Original Document:
{conversation}

Summary:
{prediction}

Fluency Evaluation Guidelines:
- Score 5: Flawless grammar, excellent readability
- Score 4: Strong fluency, minor issues
- Score 3: Acceptable fluency, some grammatical problems
- Score 2: Poor grammar, readability issues
- Score 1: Ungrammatical, difficult to understand

Question: How fluent and grammatically correct is this summary?

Score (1-5):
\end{obeylines}
\end{tcolorbox}

\paragraph{$\mathcal{P}_4$}
\textbf{System prompt:}\newline
\begin{tcolorbox}[promptbox, title=Variant 4 ($P_4$)]
\small\ttfamily\raggedright
\begin{obeylines}
You are an evaluator comparing the linguistic quality and fluency of summaries.
\end{obeylines}
\end{tcolorbox}
\textbf{Scoring prompt:}
\begin{tcolorbox}[promptbox, title=Variant 4 ($P_4$)]
\small\ttfamily\raggedright
\begin{obeylines}
Compare the fluency and grammatical quality of the generated summary.

Source:
{conversation}

Reference Summary:
{reference}

Generated Summary:
{prediction}

Rate the generated summary's fluency (grammar and readability):

1 = Ungrammatical, unreadable
2 = Poor grammar, hard to read
3 = Acceptable fluency
4 = Good fluency, minor issues
5 = Excellent fluency and grammar

Your fluency score (1-5):
\end{obeylines}
\end{tcolorbox}

\paragraph{$\mathcal{P}_5$}
\textbf{System prompt:}\newline
\begin{tcolorbox}[promptbox, title=Variant 5 ($P_5$)]
\small\ttfamily\raggedright
\begin{obeylines}
You are a fluency classifier for text summaries, rating grammatical quality on a
5-point scale.
\end{obeylines}
\end{tcolorbox}
\textbf{Scoring prompt:}
\begin{tcolorbox}[promptbox, title=Variant 5 ($P_5$)]
\small\ttfamily\raggedright
\begin{obeylines}
DOCUMENT:
{conversation}

SUMMARY:
{prediction}

TASK: Rate the summary's fluency (grammatical correctness and readability)

SCALE:
1 - Ungrammatical/Unreadable
2 - Poor Grammar
3 - Acceptable Fluency
4 - Good Grammar
5 - Excellent Fluency

FLUENCY RATING:
\end{obeylines}
\end{tcolorbox}

\subsection*{$\mathcal{G}_{13}$ \quad SummEval Relevance Judges}
\textit{Task: rate (1--5) how well the summary selects important content from the source.}

\paragraph{$\mathcal{P}_1$}
\textbf{System prompt:}\newline
\begin{tcolorbox}[promptbox, title=Variant 1 ($P_1$)]
\small\ttfamily\raggedright
\begin{obeylines}
You are an expert evaluator assessing the relevance of text summaries. Rate
summaries on a scale of 1-5 based on their selection of important content.
\end{obeylines}
\end{tcolorbox}
\textbf{Scoring prompt:}
\begin{tcolorbox}[promptbox, title=Variant 1 ($P_1$)]
\small\ttfamily\raggedright
\begin{obeylines}
You will be given one summary written for a source document.

Your task is to rate the summary on one metric: Relevance.

Please make sure you read and understand these instructions carefully. Please keep
this document open while reviewing, and refer to it as needed.

Evaluation Criteria:

Relevance (1-5) - selection of important content from the source. The summary
should include only important information from the source document. Annotators were
instructed to penalize summaries which contained redundancies and excess
information.

Evaluation Steps:

1. Read the source document carefully.
2. Read the summary and compare it to the source document.
3. Assign a score for relevance on a scale of 1-5, where 1 is the lowest and 5 is
   the highest.

Source Document:
{conversation}

Summary:
{prediction}

Evaluation Form (scores ONLY):

- Relevance (1-5):
\end{obeylines}
\end{tcolorbox}

\paragraph{$\mathcal{P}_2$}
\textbf{System prompt:}\newline
\begin{tcolorbox}[promptbox, title=Variant 2 ($P_2$)]
\small\ttfamily\raggedright
\begin{obeylines}
You are an expert at evaluating content selection. Assess how well the summary
captures important information.
\end{obeylines}
\end{tcolorbox}
\textbf{Scoring prompt:}
\begin{tcolorbox}[promptbox, title=Variant 2 ($P_2$)]
\small\ttfamily\raggedright
\begin{obeylines}
Evaluate the following summary's relevance and content selection quality.

Source Text:
{conversation}

Summary to Evaluate:
{prediction}

Assessment Criteria:
Rate the summary's relevance on a scale of 1 to 5:

5 - Perfectly captures all important content, no redundancy
4 - Captures most important content, minimal excess
3 - Captures key content but includes some irrelevant details
2 - Misses important content or includes much excess information
1 - Mostly irrelevant or redundant content

Consider:
- Does it include the most important information?
- Is there redundant or excess information?
- Are key points missing?

Your Rating (1-5):
\end{obeylines}
\end{tcolorbox}

\paragraph{$\mathcal{P}_3$}
\textbf{System prompt:}\newline
\begin{tcolorbox}[promptbox, title=Variant 3 ($P_3$)]
\small\ttfamily\raggedright
\begin{obeylines}
You are a summary quality expert focusing on content salience and importance
selection.
\end{obeylines}
\end{tcolorbox}
\textbf{Scoring prompt:}
\begin{tcolorbox}[promptbox, title=Variant 3 ($P_3$)]
\small\ttfamily\raggedright
\begin{obeylines}
Assess the relevance of this summary by evaluating its selection of important
content.

Original Document:
{conversation}

Summary:
{prediction}

Relevance Evaluation Guidelines:
- Score 5: Perfectly selects salient content, no redundancy
- Score 4: Strong content selection, captures key points
- Score 3: Adequate selection, some less important details included
- Score 2: Poor selection, misses key points or too much excess
- Score 1: Irrelevant or highly redundant content

Question: How well does this summary select important content from the source?

Score (1-5):
\end{obeylines}
\end{tcolorbox}

\paragraph{$\mathcal{P}_4$}
\textbf{System prompt:}\newline
\begin{tcolorbox}[promptbox, title=Variant 4 ($P_4$)]
\small\ttfamily\raggedright
\begin{obeylines}
You are an evaluator comparing the relevance and content selection quality of
summaries.
\end{obeylines}
\end{tcolorbox}
\textbf{Scoring prompt:}
\begin{tcolorbox}[promptbox, title=Variant 4 ($P_4$)]
\small\ttfamily\raggedright
\begin{obeylines}
Compare the relevance and content selection of the generated summary.

Source:
{conversation}

Reference Summary:
{reference}

Generated Summary:
{prediction}

Rate the generated summary's relevance (selection of important content):

1 = Mostly irrelevant or highly redundant
2 = Poor content selection
3 = Acceptable relevance
4 = Good selection of important content
5 = Excellent relevance, perfect content selection

Your relevance score (1-5):
\end{obeylines}
\end{tcolorbox}

\paragraph{$\mathcal{P}_5$}
\textbf{System prompt:}\newline
\begin{tcolorbox}[promptbox, title=Variant 5 ($P_5$)]
\small\ttfamily\raggedright
\begin{obeylines}
You are a relevance classifier for text summaries, rating content importance on a
5-point scale.
\end{obeylines}
\end{tcolorbox}
\textbf{Scoring prompt:}
\begin{tcolorbox}[promptbox, title=Variant 5 ($P_5$)]
\small\ttfamily\raggedright
\begin{obeylines}
DOCUMENT:
{conversation}

SUMMARY:
{prediction}

TASK: Rate the summary's relevance (selection of important content)

SCALE:
1 - Irrelevant/Highly Redundant
2 - Poor Content Selection
3 - Adequately Relevant
4 - Good Content Selection
5 - Perfectly Relevant

RELEVANCE RATING:
\end{obeylines}
\end{tcolorbox}

\subsection*{$\mathcal{G}_{14}$ \quad Context-Aware Judges}
\textit{Task: Likert evaluation of whether the answer uses the provided context appropriately.
Note: this group contains four judges ($\mathcal{P}_1$--$\mathcal{P}_4$).}

\paragraph{$\mathcal{P}_1$}
\textbf{System prompt:}\newline
\begin{tcolorbox}[promptbox, title=Variant 1 ($P_1$)]
\small\ttfamily\raggedright
\begin{obeylines}
You are an expert evaluator who assesses whether answers appropriately use the
provided context.
\end{obeylines}
\end{tcolorbox}
\textbf{Scoring prompt:}
\begin{tcolorbox}[promptbox, title=Variant 1 ($P_1$)]
\small\ttfamily\raggedright
\begin{obeylines}
Context Information:
{context}

Question: {conversation}

Generated Answer: {prediction}

Evaluate how well the generated answer uses the provided context:
- Does it incorporate relevant information from the context?
- Does it stay grounded in the context without hallucinating?
- Is the context properly utilized to answer the question?

Rate the quality of context usage (1-5):
1 = Ignores context or contradicts it
2 = Minimal use of context
3 = Adequate use of context
4 = Good use of context
5 = Excellent use of context

IMPORTANT: Respond with ONLY a single number (1, 2, 3, 4, or 5).
\end{obeylines}
\end{tcolorbox}

\paragraph{$\mathcal{P}_2$}
\textbf{System prompt:}\newline
\begin{tcolorbox}[promptbox, title=Variant 2 ($P_2$)]
\small\ttfamily\raggedright
\begin{obeylines}
You are a fact-checking expert who verifies if answers are grounded in the given
context.
\end{obeylines}
\end{tcolorbox}
\textbf{Scoring prompt:}
\begin{tcolorbox}[promptbox, title=Variant 2 ($P_2$)]
\small\ttfamily\raggedright
\begin{obeylines}
Given Context:
{context}

Question: {conversation}

Generated Answer: {prediction}

Task: Determine if the generated answer is factually grounded in the provided
context.
- Are all claims in the answer supported by the context?
- Is there any information that contradicts the context?
- Are there unsupported claims or hallucinations?

Answer on a scale of 1 (none what so ever) to 5 (only contains)
if it contains unsupported or contradictory information:
\end{obeylines}
\end{tcolorbox}

\paragraph{$\mathcal{P}_3$}
\textbf{System prompt:}\newline
\begin{tcolorbox}[promptbox, title=Variant 3 ($P_3$)]
\small\ttfamily\raggedright
\begin{obeylines}
You are an evaluator who assesses how well the answer demonstrates comprehension
of the provided context.
\end{obeylines}
\end{tcolorbox}
\textbf{Scoring prompt:}
\begin{tcolorbox}[promptbox, title=Variant 3 ($P_3$)]
\small\ttfamily\raggedright
\begin{obeylines}
Provided Context:
{context}

Question: {conversation}

Generated Answer: {prediction}

Evaluate how well the generated answer demonstrates comprehension of the context:
1. Does it show understanding of key information in the context?
2. Does it connect relevant pieces of information appropriately?
3. Does it answer the question based on context understanding?

Rate context comprehension (1-5):
1 = No comprehension demonstrated
2 = Poor comprehension
3 = Adequate comprehension
4 = Good comprehension
5 = Excellent comprehension

IMPORTANT: Respond with ONLY a single number (1, 2, 3, 4, or 5).
\end{obeylines}
\end{tcolorbox}

\paragraph{$\mathcal{P}_4$}
\textbf{System prompt:}\newline
\begin{tcolorbox}[promptbox, title=Variant 4 ($P_4$)]
\small\ttfamily\raggedright
\begin{obeylines}
You are an evaluator who checks if answers fully utilize all relevant parts of the
provided context.
\end{obeylines}
\end{tcolorbox}
\textbf{Scoring prompt:}
\begin{tcolorbox}[promptbox, title=Variant 4 ($P_4$)]
\small\ttfamily\raggedright
\begin{obeylines}
Available Context:
{context}

Question: {conversation}

Generated Answer: {prediction}

Evaluate if the answer uses all relevant parts of the context:
- Are there important context details that were missed?
- Does it synthesize information from multiple parts of the context when needed?
- Is the answer complete given what's available in the context?

Rate context completeness (1-5):
1 = Misses most relevant context
2 = Misses significant context details
3 = Uses some context but incomplete
4 = Uses most relevant context
5 = Thoroughly uses all relevant context

IMPORTANT: Respond with ONLY a single number (1, 2, 3, 4, or 5).
\end{obeylines}
\end{tcolorbox}

\subsection*{$\mathcal{G}_{15}$ \quad Domain-Specific Judges}
\textit{Task: rate (1--5) technical accuracy, practical applicability, or completeness of a domain answer.
Note: this group contains three judges ($\mathcal{P}_1$--$\mathcal{P}_3$) used for model training not judge agreement.}

\paragraph{$\mathcal{P}_1$}
\textbf{System prompt:}\newline
\begin{tcolorbox}[promptbox, title=Variant 1 ($P_1$)]
\small\ttfamily\raggedright
\begin{obeylines}
You are a technical expert who evaluates answers for scientific accuracy, proper
terminology, and technical precision.
\end{obeylines}
\end{tcolorbox}
\textbf{Scoring prompt:}
\begin{tcolorbox}[promptbox, title=Variant 1 ($P_1$)]
\small\ttfamily\raggedright
\begin{obeylines}
Question: {conversation}
Generated Answer: {prediction}

Evaluate this answer from a technical/scientific perspective:
- Are technical terms used correctly and precisely?
- Is the scientific information accurate and up-to-date?
- Are any formulas, processes, or methodologies correctly described?
- Is the level of technical detail appropriate for the question?

Rate the technical accuracy:
1 = Technically incorrect, major scientific errors
2 = Some technical errors, imprecise terminology
3 = Generally accurate with minor technical issues
4 = Technically sound with proper terminology
5 = Exceptionally precise and scientifically rigorous

IMPORTANT: Respond with ONLY a single number (1, 2, 3, 4, or 5). No explanations,
no additional text.
\end{obeylines}
\end{tcolorbox}

\paragraph{$\mathcal{P}_2$}
\textbf{System prompt:}\newline
\begin{tcolorbox}[promptbox, title=Variant 2 ($P_2$)]
\small\ttfamily\raggedright
\begin{obeylines}
You are a practitioner who evaluates answers based on their practical utility and
real-world applicability.
\end{obeylines}
\end{tcolorbox}
\textbf{Scoring prompt:}
\begin{tcolorbox}[promptbox, title=Variant 2 ($P_2$)]
\small\ttfamily\raggedright
\begin{obeylines}
Question: {conversation}
Generated Answer: {prediction}

Evaluate the practical value of this answer:
- Can this information be directly applied in practice?
- Are there concrete steps or actionable insights?
- Does it consider real-world constraints and limitations?
- Would a practitioner find this genuinely useful?

Rate the practical applicability:
1 = Not practical, purely theoretical with no actionable value
2 = Limited practical value, mostly abstract
3 = Some practical elements but lacks concrete guidance
4 = Practically useful with actionable insights
5 = Highly practical, immediately applicable with clear guidance

IMPORTANT: Respond with ONLY a single number (1, 2, 3, 4, or 5). No explanations,
no additional text.
\end{obeylines}
\end{tcolorbox}

\paragraph{$\mathcal{P}_3$}
\textbf{System prompt:}\newline
\begin{tcolorbox}[promptbox, title=Variant 3 ($P_3$)]
\small\ttfamily\raggedright
\begin{obeylines}
You are an evaluator who assesses whether answers comprehensively address the
question's scope and complexity.
\end{obeylines}
\end{tcolorbox}
\textbf{Scoring prompt:}
\begin{tcolorbox}[promptbox, title=Variant 3 ($P_3$)]
\small\ttfamily\raggedright
\begin{obeylines}
Question: {conversation}
Generated Answer: {prediction}

Evaluate the completeness and scope coverage:
- Does the answer address all parts of the question?
- Are important related concepts or considerations included?
- Is the breadth of coverage appropriate for the question's complexity?
- Are there significant omissions or gaps?

Rate the completeness:
1 = Major gaps, incomplete coverage of the question
2 = Some important aspects missing
3 = Adequate coverage with minor omissions
4 = Comprehensive coverage of most relevant aspects
5 = Exceptionally complete, addresses all relevant dimensions

IMPORTANT: Respond with ONLY a single number (1, 2, 3, 4, or 5). No explanations,
no additional text.
\end{obeylines}
\end{tcolorbox}

\vspace{0.3cm}


\newpage

\end{document}